\documentclass[final,5p,times,twocolumn]{elsarticle}

\usepackage{amsthm,amsmath,amssymb}
\usepackage{mathrsfs}
\usepackage{booktabs}
\usepackage{multirow}
\usepackage{multicol}
\usepackage{color}
\usepackage{hyperref}
\usepackage{float}
\usepackage{subfig}

\usepackage[table]{xcolor} 
\usepackage{colortbl} 
\usepackage{rotating} 
\usepackage{graphicx}

\newcommand{\tabincell}[2]{\begin{tabular}{@{}#1@{}}#2\end{tabular}}
\definecolor{LightCyan}{rgb}{0.88,1,1}
\newcommand{\gray}[1]{\textcolor{gray}{#1}}

\journal{Elsevier}

\begin{document}

\begin{frontmatter}



\title{HiCMAE: Hierarchical Contrastive Masked Autoencoder for Self-Supervised Audio-Visual Emotion Recognition}

\author[label1,label2]{Licai Sun}\ead{sunlicai2019@ia.ac.cn}
\author[label1]{Zheng Lian}\ead{lianzheng2016@ia.ac.cn}
\author[label1,label2]{Bin Liu\corref{cor}}\ead{liubin@nlpr.ia.ac.cn}
\author[label3,label4]{Jianhua Tao\corref{cor}}\ead{jhtao@tsinghua.edu.cn}
\cortext[cor]{Corresponding authors: Biu Liu and Jianhua Tao.}

\affiliation[label1]{organization={School of Artificial Intelligence, University of Chinese Academy of Sciences},
            city={Beijing},
            country={China}}

\affiliation[label2]{organization={Institute of Automation, Chinese Academy of Sciences},
            city={Beijing},
            country={China}}
\affiliation[label3]{organization={Department of Automation, Tsinghua University},
            city={Beijing},
            country={China}}
\affiliation[label4]{organization={Beijing National Research Center for Information Science and Technology, Tsinghua University},
            city={Beijing},
            country={China}}



\begin{abstract}
Audio-Visual Emotion Recognition (AVER) has garnered increasing attention in recent years for its critical role in creating emotion-aware intelligent machines. Previous efforts in this area are dominated by the supervised learning paradigm. Despite significant progress, supervised learning is meeting its bottleneck due to the longstanding data scarcity issue in AVER. 
Motivated by recent advances in self-supervised learning, we propose Hierarchical Contrastive Masked Autoencoder (HiCMAE), a novel self-supervised framework that leverages large-scale self-supervised pre-training on vast unlabeled audio-visual data to promote the advancement of AVER. 
Following prior arts in self-supervised audio-visual representation learning, HiCMAE adopts two primary forms of self-supervision for pre-training, namely masked data modeling and contrastive learning. 
Unlike them which focus exclusively on top-layer representations while neglecting explicit guidance of intermediate layers, HiCMAE develops a \textit{three-pronged} strategy to foster \textit{hierarchical} audio-visual feature learning and improve the overall quality of learned representations.
Firstly, it incorporates \textit{hierarchical skip connections} between the encoder and decoder to encourage intermediate layers to learn more meaningful representations and bolster masked audio-visual reconstruction. 
Secondly, \textit{hierarchical cross-modal contrastive learning} is also exerted on intermediate representations to narrow the audio-visual modality gap progressively and facilitate subsequent cross-modal fusion. 
Finally, during downstream fine-tuning, HiCMAE employs \textit{hierarchical feature fusion} to comprehensively integrate multi-level features from different layers. 
To verify the effectiveness of HiCMAE, we conduct extensive experiments on 9 datasets covering both categorical and dimensional AVER tasks. Experimental results show that our method significantly outperforms state-of-the-art supervised and self-supervised audio-visual methods, which indicates that HiCMAE is a powerful audio-visual emotion representation learner. 
Codes and models are publicly available at \textcolor[rgb]{0.93,0.0,0.47}{\url{https://github.com/sunlicai/HiCMAE}}.

\end{abstract}




\end{frontmatter}



\section{Introduction}

\begin{quote}
\emph{“The question is not whether intelligent machines can have any emotions, but whether machines can be intelligent without any emotions.”} 
\begin{flushright}
\textemdash Marvin Minsky
\end{flushright}
\end{quote}

\begin{figure}[]
	\centering
    \includegraphics[width=1.0\linewidth]{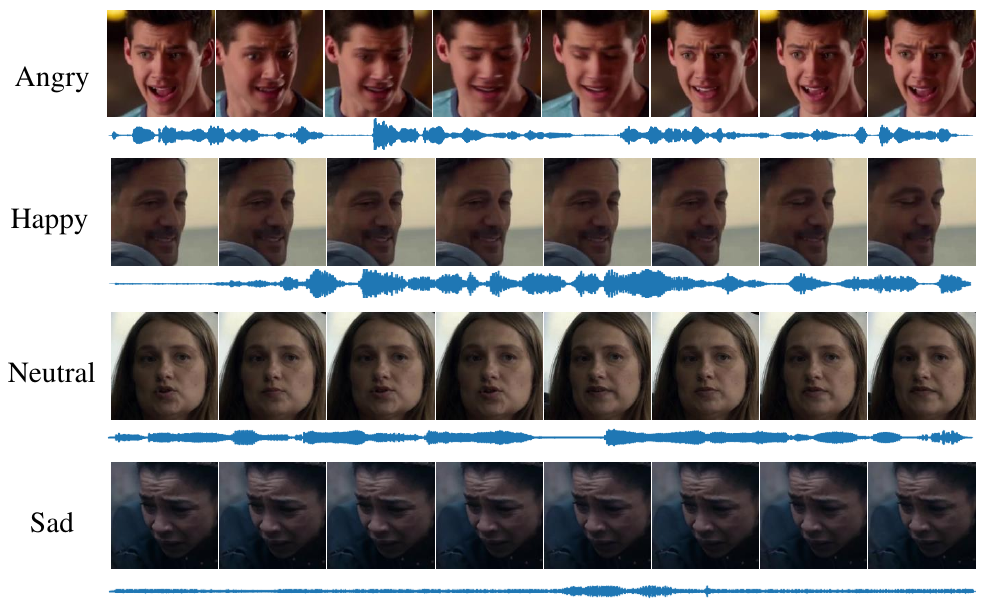}
    \caption{Several samples selected from the MAFW \cite{liu2022mafw} dataset.}
	\label{fig_aver_example}
\end{figure}

Emotions are fundamental to the multifaceted spectrum of human experience, influencing our cognition, decision-making, and interpersonal interactions \cite{schwarz2000emotion}. They also play a pivotal role in developing intelligent machines and achieving the ultimate goal of emotional artificial intelligence, as Marvin Minsky, a pioneer of artificial intelligence, highlighted above \cite{minsky1988society}. 
Typically, people express and perceive emotions through multiple modalities. Previous psychological studies have demonstrated that the language modality (i.e., verbal information) only contributes 7\% to the perception of emotions in our daily communication, while the audio (e.g., tone and intonation) and visual (e.g., facial expressions) modalities significantly predominate, contributing to 38\% and 55\% respectively \cite{mehrabian1968communication, wu2014survey}. As a result, the last two decades have witnessed a surge of interest in automatic Audio-Visual Emotion Recognition (AVER) in the affective computing community. 
The task of AVER is to extract emotion-related representations from audio-visual signals and then integrate them to identify the subject's emotional state. Several concrete examples are depicted in Fig. \ref{fig_aver_example}.

Early studies on AVER concentrate on developing various hand-engineering features for audio and video modalities \cite{zeng2008survey, wu2014survey}.
With the advent of the deep learning era, a new trend has emerged towards learning features directly from raw audio-visual data by training deep supervised neural networks in an end-to-end manner \cite{zhang2017learning, tzirakis2017end, zhang2023transformer}.
Despite considerable advancements, supervised learning is heavily constrained by its reliance on large amounts of labeled data to achieve satisfactory performance. This reliance significantly hampers further progress of supervised methods due to the longstanding issue of data scarcity in AVER \cite{zhang2023deep,pei2024affective}.

Recently, another deep learning paradigm, i.e., self-supervised learning, which can learn powerful representations from vast unlabeled data, has revolutionized many research areas \cite{balestriero2023cookbook, devlin2018bert, mao2022biases, li2023skier, radford2021learning, he2022masked}. In particular, masked data modeling (e.g., MAE \cite{he2022masked}) and contrastive learning (e.g., CLIP \cite{radford2021learning}) are two prominent methods and have demonstrated great success in self-supervised visual representation learning. Specifically, masked data modeling aims to reconstruct the raw data from masked inputs, while contrastive learning encourages the semantic alignment of different modalities. They are also combined and extended to learn generic audio-visual representations in recent studies, such as CAV-MAE \cite{gong2023contrastive} and MAViL \cite{huang2023mavil}. However, due to the domain gap between upstream pre-training and downstream AVER tasks, the learned representations are typically not suitable for AVER. More recently, Sadok et al. \cite{sadok2023vector} developed a vector-quantized masked autoencoder (VQ-MAE-AV) for AVER. Although achieving promising results, VQ-MAE-AV requires pre-trained variational autoencoders for vector quantization and thus cannot be pre-trained in a single stage. 
More importantly, several studies on masked image modeling have shown that intermediate layers are essential to self-supervised visual representation learning and explicitly guiding them can improve the quality of learned representations \cite{wang2023masked, liu2023improving}. However, the aforementioned methods fail to achieve that goal as they solely operate on representations from the top layer and neglect explicit guidance on intermediate layers, thus impeding feature evolution across layers and leading to sub-optimal results in downstream tasks.

\begin{figure}[t]
	\centering
    \includegraphics[width=1.0\linewidth]{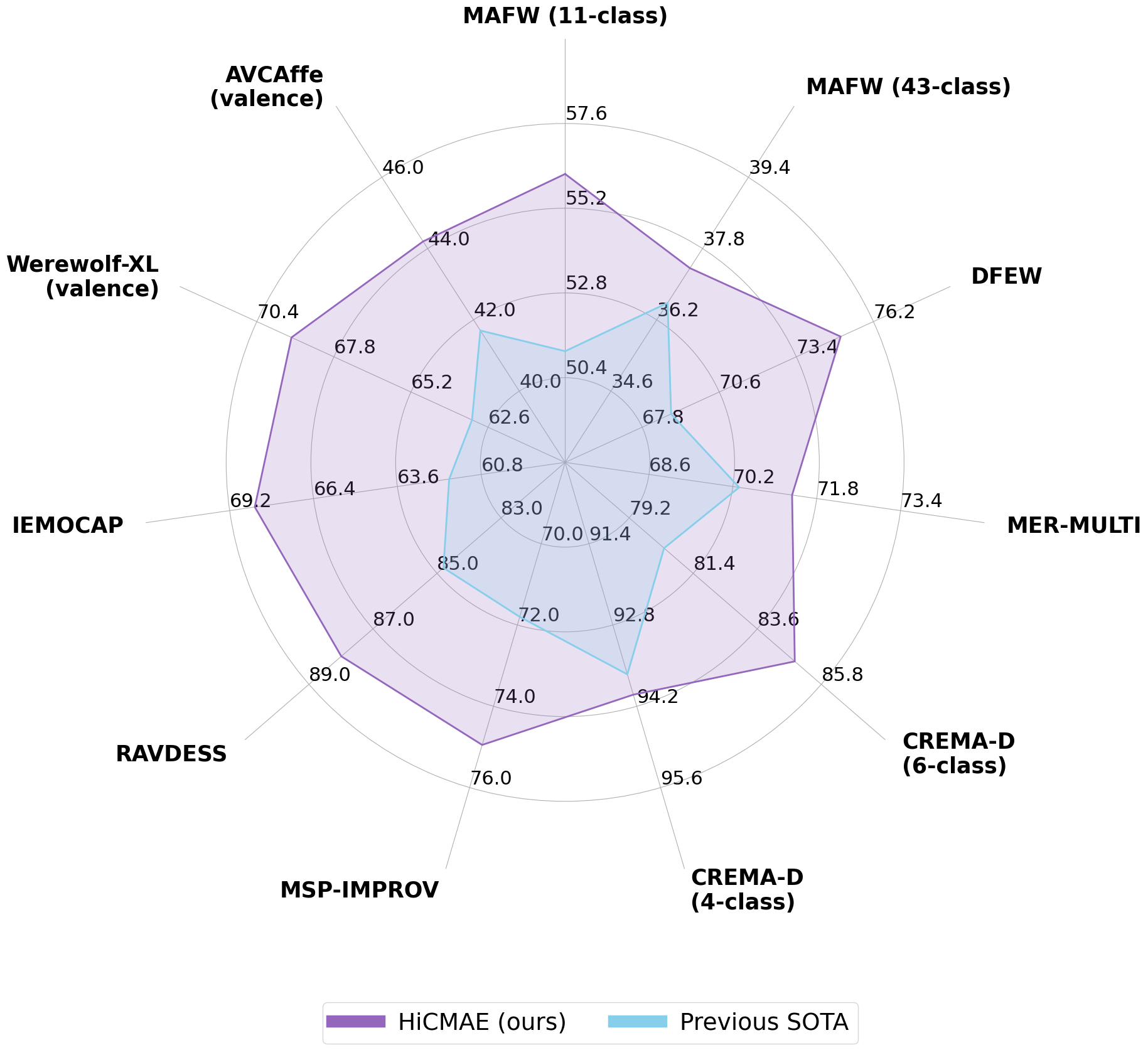}
    \caption{Comparison with state-of-the-art audio-visual methods on 9 datasets. We present Pearson correlation coefficient on Werewolf-XL \cite{zhang2021werewolf} and weighted F1-score on AVCAffe \cite{sarkar2022avcaffe} and MER-MULTI \cite{lian2023mer}. For other datasets, we show weighted average recall (WAR).}
	\label{fig_sota_comparision_radar_plot}
\end{figure}

To address the above challenges, this paper builds on top of MAE and contrastive learning to propose Hierarchical Contrastive Masked Autoencoder (HiCMAE), a novel self-supervised framework tailored for AVER. As shown in Fig. \ref{fig_overview}, it is mainly composed of two audio-visual encoders, a cross-modal fusion encoder, and two lightweight audio-visual decoders. 
To foster hierarchical audio-visual feature learning and improve the overall quality of learned representations, HiCMAE develops a \textit{three-pronged} strategy:
(1) Drawing inspiration from the architecture design in U-Net \cite{ronneberger2015u}, HiCMAE introduces \textit{hierarchical skip connections} between the encoder and decoder to drive intermediate encoder layers to learn more useful representations and aid the decoder in accomplishing the task of masked audio-visual reconstruction. (2) Considering the natural audio-visual correspondences in videos, \textit{hierarchical cross-modal contrastive learning} is also applied to intermediate representations of audio-visual encoders to reduce heterogeneous modality gap in a progressive manner and enhance cross-modal fusion in subsequent layers. (3) Since different layers typically capture distinct levels of information, HiCMAE performs \textit{hierarchical feature fusion} during downstream fine-tuning to comprehensively integrate multi-level features from various encoder layers. 
To demonstrate the effectiveness of HiCMAE, we perform large-scale self-supervised pre-training on VoxCeleb2 \cite{chung2018voxceleb2} and evaluate the pre-trained model on nine datasets encompassing both categorical and dimensional AVER tasks. As illustrated in Fig. \ref{fig_sota_comparision_radar_plot}, our HiCMAE significantly outperforms state-of-the-art supervised or self-supervised audio-visual methods. 
For example, HiCMAE surpasses the best-performing VQ-MAE-AV \cite{sadok2023vector} by \textbf{+4.49\%} WAR on CREMA-D (6-class) \cite{cao2014crema} and beats the state-of-the-art T-MEP \cite{zhang2023transformer} by \textbf{+5.02\%} WAR on MAFW (11-class) \cite{liu2022mafw} and \textbf{+6.16\%} WAR on DFEW \cite{jiang2020dfew}. 
Moreover, extensive ablation studies also justify various design choices in HiCMAE.

In summary, our main contributions are three-fold:
\begin{itemize}
\item We present HiCMAE, a novel self-supervised framework for AVER, as an early endeavor to leverage large-scale self-supervised pre-training to address the dilemma of supervised methods and promote the development of AVER. 
\item Unlike previous methods, HiCMAE introduces a three-pronged approach to foster hierarchical audio-visual feature learning and the ablation studies verify its efficacy.
\item Comprehensive experiments across 9 datasets covering both categorical and dimensional AVER tasks demonstrate that HiCMAE beats state-of-the-art audio-visual methods by significant margins, indicating that \textbf{HiCMAE is a powerful audio-visual emotion representation learner}. 
\end{itemize}

\section{Related Work}

\subsection{Audio-Visual Emotion Recognition}
Most studies on Audio-Visual Emotion Recognition (AVER) fall into the supervised learning paradigm. They mainly focus on two important aspects: unimodal feature extraction and audio-visual information fusion. 
As for the first aspect, researchers have developed and exploited numerous features in the past two decades \cite{zeng2008survey, wu2014survey, zhang2023deep, lian2024merbench}. Early studies concentrate on various handcrafted features for two modalities, such as IS13 \cite{schuller2013interspeech} and eGeMAPS \cite{eyben2015geneva} for audio, LBP-TOP \cite{zhao2007dynamic} and HOG \cite{dalal2005histograms} for video. 
With the advent of deep learning, a large amount of deep supervised models trained on large audio and image/video datasets have emerged as powerful audio-visual feature extractors \cite{fan2016video, chen2017multimodal, sun2020multi, sun2021multimodal, meng2022valence, lian2023mer}, such as PANNs \cite{kong2020panns} and VGGish \cite{hershey2017cnn} for audio, VGGFace \cite{cao2018vggface2} and C3D \cite{tran2015learning} for video. 
There are also lots of attempts to train end-to-end deep neural networks on raw audio and video emotion data \cite{trigeorgis2016adieu, tzirakis2017end, huang2018end, jiang2020dfew, wang2022ferv39k}.
In recent years, plenty of large self-supervised pre-trained models have demonstrated great success in audio (e.g., Wav2vec2.0 \cite{baevski2020wav2vec} and HuBERT \cite{hsu2021hubert}) or video (e.g., SVFAP \cite{sun2023svfap} and MAE-DFER \cite{sun2023mae}) emotion recognition.
After unimodal feature extraction, the next crucial step is audio-visual information fusion. Current fusion strategies can be roughly divided into three categories, i.e., early fusion, late fusion, and model-level fusion \cite{zeng2008survey, wu2014survey}. Early fusion typically combines audio-visual features at the input level \cite{chen2017multimodal, meng2022valence}. In contrast, late fusion integrates audio-visual predictions at the decision level \cite{sun2020multi, sun2021multimodal}. The most commonly used strategy is model-level fusion \cite{zhang2017learning, tsai2019multimodal, huang2020multimodal, goncalves2022auxformer, lian2021ctnet, tran2022pre, hsu2023applying, sun2023efficient, zhang2023transformer}. For example, MulT \cite{tsai2019multimodal} utilizes cross-modal attention to capture dense interactions in unaligned audio-visual feature sequences. EMT \cite{sun2023efficient} improves MulT by introducing the global multimodal context and employing it to interact with local unimodal features to achieve efficient cross-modal information exchange. The recent T-MEP \cite{zhang2023transformer} adopts a similar strategy with MulT to fuse fine-grained audio-visual tokens by combining self-attention and cross-attention mechanisms in an interleaving manner.

Despite promising results, the aforementioned methods are mostly supervised learning methods. They either train from scratch or rely on pre-trained models from other different tasks for initialization, which are thus severely constrained by the data scarcity issue in AVER or suffer from substantial domain shifts. 
In contrast, this paper presents an early attempt to leverage large-scale self-supervised audio-visual pre-training on massive unlabeled data to largely advance the development of audio-visual emotion recognition.

\subsection{Self-Supervised Audio-Visual Representation Learning}
In general, there are three kinds of methods for learning generic audio-visual representations through self-supervision, namely contrastive learning, masked data modeling, and the hybrid method. Contrastive learning exploits the natural audio-visual correspondence in videos as a free signal for self-supervision \cite{arandjelovic2017look, owens2018audio, morgado2021audio}. In contrast, the goal of masked data modeling is to reconstruct the original data from its masked input. Motivated by its recent success in the image and video domain \cite{he2022masked, tong2022videomae, feichtenhofer2022masked}, Georgescu et al. \cite{georgescu2023audiovisual} extend it to the audio-visual domain and achieve significant improvements over previous methods. Recently, a few studies have made attempts to combine the former two kinds of methods, resulting in the hybrid method. In particular, CAV-MAE \cite{gong2023contrastive} integrates MAE and cross-modal contrastive learning and shows that they are complementary. MAViL \cite{huang2023mavil} further introduces intra-modal contrastive learning and masked self-training on contextualized features to improve audio-visual pre-training. 
Although these methods have demonstrated great success in general audio-visual tasks, the learned representations are typically not suitable for AVER because they are trained on general scene or action videos instead of facial videos in AVER. 
Recently, VQ-MAE-AV \cite{sadok2023vector} introduces a vector-quantized MAE for AVER. Despite promising results, it requires two-stage pre-training. 
In addition, the more important issue is that the aforementioned audio-visual masked autoencoders fail to promote hierarchical audio-visual feature learning as they focus exclusively on top-layer representations while neglecting explicit guidance of intermediate layers, thus impeding feature evolution across layers and leading to sub-optimal performance \cite{wang2023masked, liu2023improving}.  
Therefore, our HiCAME presents a three-pronged approach to foster hierarchical audio-visual feature learning and demonstrates improved performance.

\section{Method}
\begin{figure*}[t]
	\centering
    \includegraphics[width=1.0\linewidth]{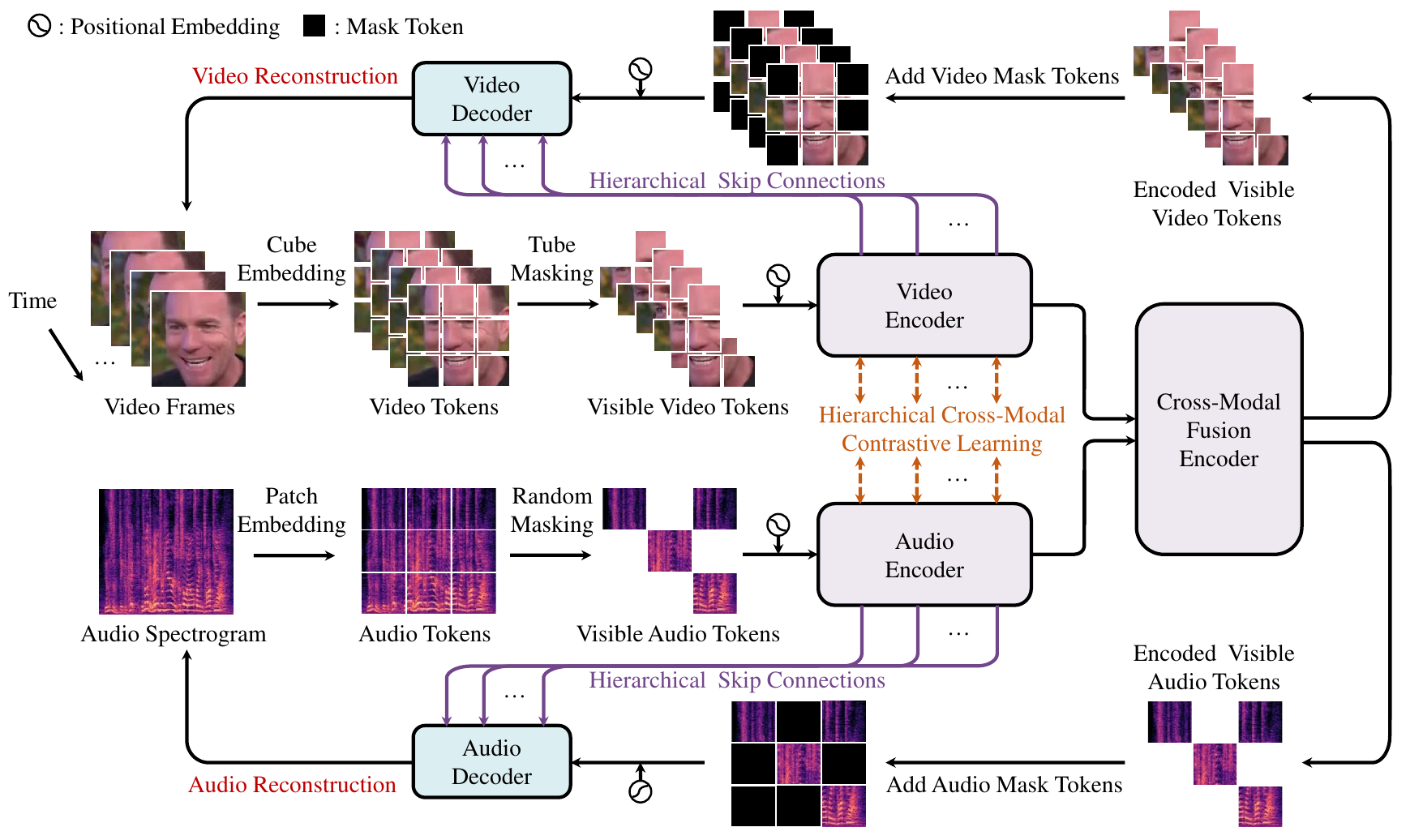}
    \caption{The overall pre-training pipeline of HiCMAE. HiCMAE mainly adopts an asymmetric encoder-decoder architecture with hierarchical skip connections in between for masked audio-visual reconstruction. Besides, hierarchical cross-modal contrastive learning is employed at intermediate audio-visual encoder layers to reduce the modality gap in a progressive manner and facilitate cross-modal fusion in subsequent layers.}
	\label{fig_overview}
\end{figure*}
In this section, we elaborate on Hierarchical Contrastive Masked AutoEncoder (HiCMAE), a novel self-supervised audio-visual emotion representation learning framework for AVER.
The training process of HiCMAE includes two steps, i.e., self-supervised pre-training (Section \ref{sec_method_pretrain_masked_data_modeling}-\ref{sec_method_pretrain_loss}) on large-scale unlabeled AVER data and downstream fine-tuning (Section \ref{sec_method_finetuning}) on limited labeled AVER data.
Specifically, the self-supervised pre-training pipeline of HiCMAE is illustrated in Fig.~\ref{fig_overview}. 
It mainly consists of two modality-specific encoders, a cross-modal fusion encoder, and two lightweight modality-specific decoders. 
HiCMAE adopts two primary forms of self-supervision for pre-training: masked data modeling (i.e., masked audio-visual reconstruction) and contrastive learning. 
Moreover, it introduces a three-pronged strategy to promote hierarchical audio-visual feature learning during both pre-training and fine-tuning, including hierarchical skip connections between the encoder and decoder (Section \ref{sec_method_pretrain_masked_data_modeling}), hierarchical cross-modal contrastive learning (Section \ref{sec_method_pretrain_contrastive_learning}), and hierarchical feature fusion for downstream fine-tuning (Section \ref{sec_method_finetuning}).

\subsection{Masked Audio-Visual Reconstruction with Hierarchical Skip Connections}
\label{sec_method_pretrain_masked_data_modeling}
As depicted in Fig. \ref{fig_overview}, HiCMAE follows MAE \cite{he2022masked} to adopt an asymmetric encoder-decoder architecture for efficient self-supervised audio-visual pre-training. The raw audio and video inputs are first embedded into tokens and then masked to filter out a substantial proportion of tokens. Next, the visible (i.e., unmasked) audio and video tokens are processed by their modality-specific encoders and a cross-modal fusion encoder. After feature encoding, the visible audio and video tokens are padded with learnable masked tokens and then passed through lightweight modality-specific decoders for final audio and video reconstruction. Note that, unlike conventional masked autoencoders, hierarchical skip connections are added between audio-visual encoders and decoders to better guide the encoder feature learning at different levels and promote masked audio-visual reconstruction by providing the decoder with multi-level features.

\subsubsection{Data Embedding and Token Masking}
Since we use Transformer \cite{vaswani2017attention} as the main component of the encoder, it is necessary to embed audio and video inputs into a sequence of discrete tokens. 
Formally, we denote the facial video frames as $\mathbf{X}_v \in \mathbb{R}^{T_v\times H \times W \times 3}$ ($T_v$ is the number of frames, $H$ and $W$ denote the height and width, and $3$ represents RGB channels) and the audio spectrogram as $\mathbf{X}_a \in \mathbb{R}^{T_a \times F}$ ($T_a$ is the temporal length and $F$ denotes the number of frequency channels). 
We utilize a cube embedding layer with a size of $2\times 16 \times 16$ to split $\mathbf{X}_v$ into video tokens 
$\mathbf{X}_v^{'} \in \mathbb{R}^{N_v \times C}$ ($N_v = \frac{T}{2} \cdot \frac{H}{16} \cdot \frac{W}{16}$ is the number of video tokens, $C$ is the number of feature channels). As for $\mathbf{X}_a$, we employ a patch embedding layer with a size of $16 \times 16$ to split it into audio tokens $\mathbf{X}_a^{'} \in \mathbb{R}^{N_a \times C}$ ($N_a = \frac{T_a}{16} \cdot \frac{F}{16}$ is the number of audio tokens).

After data embedding, we mask out a large proportion of audio and video tokens to make audio-visual reconstruction a non-trivial self-supervised task and significantly reduce the pre-training cost at the same time. Considering the high temporal redundancy and correlation in video data, we adopt the tube masking (i.e., each temporal slice has the same masking pattern) strategy \cite{tong2022videomae}. For audio, we simply use random masking \cite{huang2022masked}. The masking ratios for audio and video tokens are set to $\rho_a = 80\%$ and $\rho_v = 90\%$ respectively \cite{tong2022videomae, huang2022masked}. After token masking, only the visible video tokens $\mathbf{X}_v^{''} \in \mathbb{R}^{N_v' \times C}$ ($N_v' = (1 - \rho_v) \cdot N_v$) and audio tokens $\mathbf{X}_a^{''} \in \mathbb{R}^{N_a' \times C}$ ($N_a' = (1 - \rho_a) \cdot N_a$) will be processed by the encoder introduced next. 

\subsubsection{Encoder}
The encoder in HiCMAE comprises two modality-specific encoders and a cross-modal fusion encoder. The former respects the diversity of audio-visual information and aims to learn unique characteristics in each modality, while the latter is used to capture meaningful cross-modal interactions and reinforce the representation of one modality with the supplementary information from the other modality for better masked audio-visual reconstruction.

\textbf{Modality-Specific Encoder.} As a general self-supervised pre-training framework, there can be many choices for the architecture of modality-specific encoders. For simplicity, we follow MAE \cite{he2022masked} and VideoMAE \cite{tong2022videomae} to employ standard Transformer \cite{vaswani2017attention} as audio and video encoders. Other efficient architectures such as LGI-Former \cite{sun2023mae} can be explored in future work. 
The audio and video encoders consist of $N_s$ Transformer layers. Each Transformer layer is mainly composed of Multi-Head Self-Attention (MHSA) and Feed-Forward Network (FFN):
\begin{equation}
\begin{split}
\mathbf{E}^{j-1'}_m &= \textrm{MHSA}(\textrm{LN}(\mathbf{E}^{j-1}_m)) + \mathbf{E}^{j-1}_m, \\
\mathbf{E}^{j}_m &= \textrm{FFN}(\textrm{LN}(\mathbf{E}^{j-1'}_m)) + \mathbf{E}^{j-1'}_m,
\end{split}
\label{eq_transformer}
\end{equation}
where $\mathbf{E}^{0}_m = \mathbf{X}_m^{''}$,  $m \in \{a,v\}$ denotes the audio or video modality, $j \in \{1,..., N_s\}$ is the layer index, and $\textrm{LN}$ stands for layer normalization \cite{vaswani2017attention}. For MHSA in Eq. (\ref{eq_transformer}), it calculates dot-product attention to learn intrinsic dependency relationships in input audio/video tokens: 
\begin{equation}
\begin{split}
\textrm{MHSA}(\mathbf{E}) &= \textrm{Concat}(\textrm{head}_{1}, ..., \textrm{head}_{H})\mathbf{W}^O, \\
\textrm{head}_{h} &= \textrm{Softmax}(\frac{\mathbf{Q}_h \mathbf{K}_h^\top}{\sqrt{d_h}})\mathbf{V}_h, h=1,..., H, \\
\mathbf{Q}_h &= \mathbf{E} \mathbf{W}^{Q}_h, \mathbf{K}_h = \mathbf{E}  \mathbf{W}^{K}_h, \mathbf{V}_h = \mathbf{E} \mathbf{W}^{V}_h,
\end{split}
\label{eq_mhsa}
\end{equation}
where $\mathbf{W}^Q_h \in \mathbb{R}^{C\times d_h}$, $\mathbf{W}^K_h \in \mathbb{R}^{C\times d_h}$, $\mathbf{W}^V_h \in \mathbb{R}^{C\times d_h}$, $\mathbf{W}^{O} \in \mathbb{R}^{C\times C}$, $H$ is the number of attention heads, and $d_h=C/H$ is the feature dimension in each attention head.
For FFN in Eq. (\ref{eq_transformer}), it comprises two linear projection layers with a GELU \cite{hendrycks2016gaussian} activation function in between:
\begin{equation}
\textrm{FFN}(\mathbf{E}) = \textrm{GELU}(\mathbf{E} \mathbf{W}_1+\mathbf{b}_1) \mathbf{W}_2 + \mathbf{b}_2,
\label{eq_ffn}
\end{equation}
where $\mathbf{W}_1 \in \mathbb{R}^{C\times 4C}$, $\mathbf{b}_1 \in \mathbb{R}^{4C}$, $\mathbf{W}_2 \in \mathbb{R}^{4C\times C}$, and $\mathbf{b}_2 \in \mathbb{R}^{C}$ are learnable parameters.

\textbf{Cross-Modal Fusion Encoder.}
After unimodal feature encoding, HiCMAE employs a cross-modal fusion encoder to capture meaningful interactions in audio-visual modalities and enable cross-modal reinforcement \cite{tsai2019multimodal, sun2023efficient}. We mainly utilize multi-head cross-attention (MHCA) to implement this module. MHCA shares the same spirit with MHSA. The main difference is that MHCA accepts two modalities as input, with one as the target modality and the other as the source modality. The formulation of MHCA is given as follows:
\begin{equation}
\begin{split}
\textrm{MHCA}(\mathbf{E}, \mathbf{F}) &= \textrm{Concat}(\textrm{head}_{1}, ..., \textrm{head}_{H})\mathbf{W}^O, \\
\textrm{head}_{h} &= \textrm{Softmax}(\frac{\mathbf{Q}_h^E {\mathbf{K}_h^F}^\top}{\sqrt{d_h}})\mathbf{V}_h^F, h=1,...,H, \\
\mathbf{Q}_h^E &= \mathbf{E} \mathbf{W}^{Q}_h, \mathbf{K}_h^F = \mathbf{F}  \mathbf{W}^{K}_h, \mathbf{V}_h^F = \mathbf{F} \mathbf{W}^{V}_h,
\end{split}
\label{eq_mhca}
\end{equation}
where the notation is similar to that in Eq. (\ref{eq_mhsa}).
By combining MHCA, MHSA, and FFN, we obtain one half of the fusion encoder which can achieve cross-modal reinforcement from \textit{audio} to \textit{video}:
\begin{equation}
\begin{split}
\mathbf{E}^{j'}_{a \rightarrow v}  &= \textrm{MHCA}(\textrm{LN}(\mathbf{E}^{j-1}_{a \rightarrow v}), \textrm{LN}(\mathbf{E}^{j-1}_{v \rightarrow a})) + \mathbf{E}^{j-1}_{a \rightarrow v}, \\
\mathbf{E}^{j''}_{a \rightarrow v}  &= \textrm{MHSA}(\textrm{LN}(\mathbf{E}^{j'}_{a \rightarrow v})) + \mathbf{E}^{j'}_{a \rightarrow v}, \\
\mathbf{E}^{j}_{a \rightarrow v}  &= \textrm{FFN}(\textrm{LN}(\mathbf{E}^{j''}_{a \rightarrow v})) + \mathbf{E}^{j''}_{a \rightarrow v},
\end{split}
\label{eq_fusion_a2v}
\end{equation}
where $j \in \{1, ..., N_f\}$ is the layer index, $N_f$ is the number of fusion encoder layers, $\mathbf{E}^{0}_{a \rightarrow v} = \mathbf{E}^{N_s}_{v}$ is the output of video encoder, and $\mathbf{E}^{0}_{v \rightarrow a} = \mathbf{E}^{N_s}_{a}$ is the output of audio encoder.
Similarly, the other half of the fusion encoder enables cross-modal reinforcement from \textit{video} to \textit{audio}:
\begin{equation}
\begin{split}
\mathbf{E}^{j'}_{v \rightarrow a}  &= \textrm{MHCA}(\textrm{LN}(\mathbf{E}^{j-1}_{v \rightarrow a}), \textrm{LN}(\mathbf{E}^{j-1}_{a \rightarrow v})) + \mathbf{E}^{j-1}_{v \rightarrow a}, \\
\mathbf{E}^{j''}_{v \rightarrow a}  &= \textrm{MHSA}(\textrm{LN}(\mathbf{E}^{j'}_{v \rightarrow a})) + \mathbf{E}^{j'}_{v \rightarrow a}, \\
\mathbf{E}^{j}_{v \rightarrow a}  &= \textrm{FFN}(\textrm{LN}(\mathbf{E}^{j''}_{v \rightarrow a})) + \mathbf{E}^{j''}_{v \rightarrow a}.
\end{split}
\label{eq_fusion_v2a}
\end{equation}

\subsubsection{Decoder}
\label{section_decoder}
After feature encoding, HiCMAE utilizes two modality-specific decoders for final masked audio-visual reconstruction. Previous studies have demonstrated that a small-capacity model is enough to reconstruct masked information \cite{he2022masked, tong2022videomae, huang2022masked}. Therefore, we follow them to employ a lightweight Transformer-based model as the decoder, whose number of layers is much less than that of the encoder, to largely reduce the computation burden during self-supervised pre-training. 

To begin with, we denote audio and video tokens output by the last fusion encoder layer as $\mathbf{E}^{N_f}_{v \rightarrow a}$ and $\mathbf{E}^{N_f}_{a \rightarrow v}$ respectively. Then we pad them with learnable mask tokens $\mathbf{M}_m$ ($m \in \{a,v\}$), and fixed sinusoidal positional embeddings $\mathbf{PE}_d$ are added to retain positional information. After that, we obtain the input tokens for audio and video decoders, i.e., $\mathbf{D}^{0}_a = [\mathbf{E}^{N_f}_{v \rightarrow a}, \mathbf{M}_a] + \mathbf{PE}_d$ and $\mathbf{D}^{0}_a = [\mathbf{E}^{N_f}_{a \rightarrow v}, \mathbf{M}_v] + \mathbf{PE}_d$.

\begin{figure}[t]
	\centering
    \includegraphics[width=1.0\linewidth]{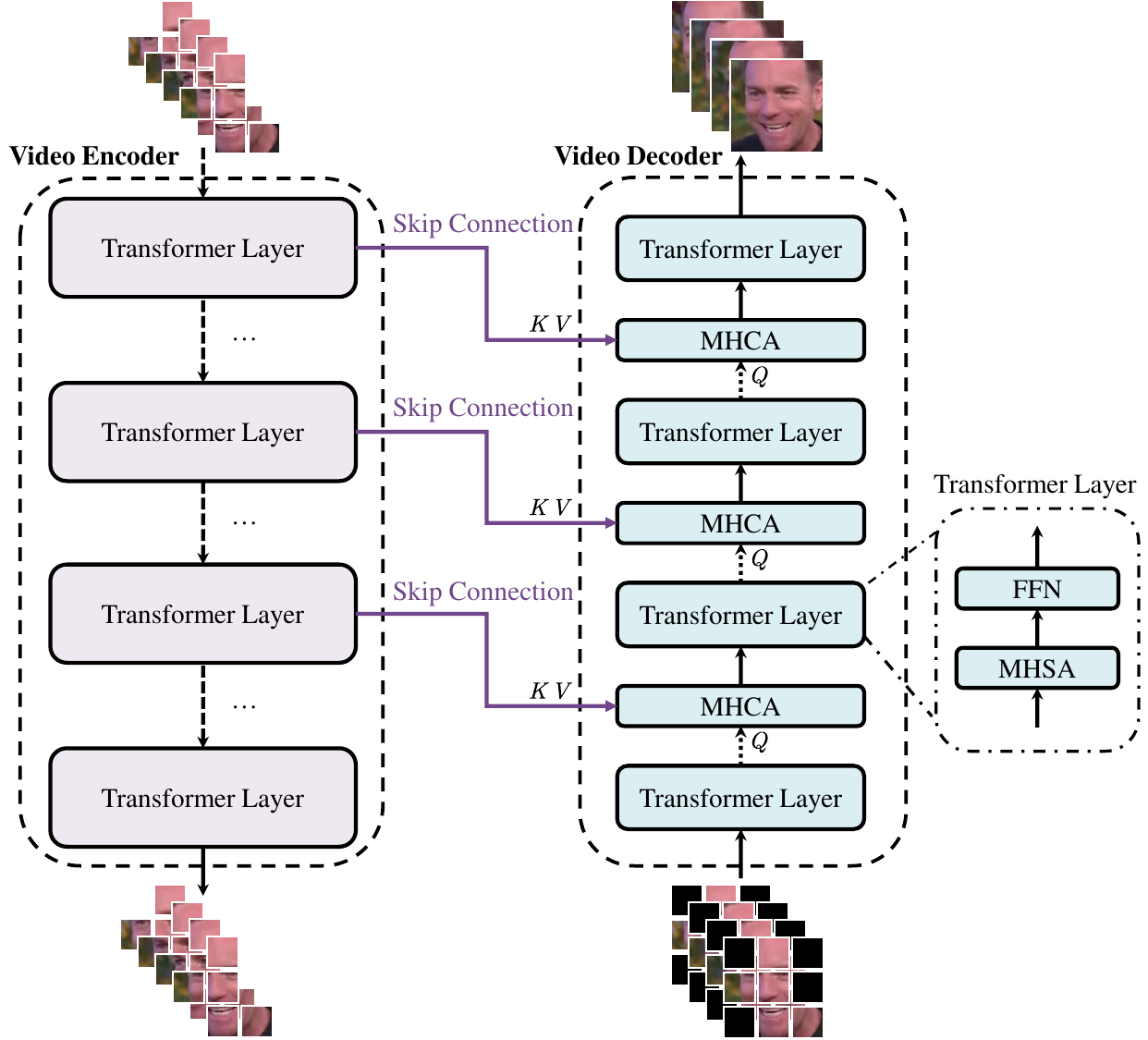}
    \caption{The illustration of hierarchical skip connections between the encoder and decoder (taking the video modality as an example).}
	\label{fig_hierarchical_skip_connections}
\end{figure}
\textbf{Hierarchical Skip Connections}. As shown in Fig. \ref{fig_hierarchical_skip_connections}, unlike previous audio-visual masked autoencoders which \textit{solely} operate on encoder representation from the last layer and neglect explicit guidance on other layers \cite{gong2023contrastive, georgescu2023audiovisual, huang2023mavil, sadok2023vector}, HiCMAE incorporates hierarchical skip connections between intermediate encoder and decoder layers to explicitly steer encoder feature learning of different levels and assist the decoder to complete the task of masked audio-visual reconstruction. 
Specifically, HiCMAE adds an MHCA layer before each (except for the first) Transformer layer in the decoder. 
Given the tokens $\mathbf{D}^k_m$ ($m \in \{a,v\}$) output by the $k$th Transformer layer, the skip connection constructed by MHCA enables it to directly attend to the intermediate encoder representations $\mathbf{E}^j_m$ with different levels of information (from high-level coarse-grained to low-level fine-grained) and extract useful information to reinforce itself. Then, as usual, a standard Transformer layer is followed to exploit global self-attention to infer masked information. The whole process can be formulated as follows:
\begin{equation}
\begin{split}
\mathbf{D}^{k'}_{m}  &= \textrm{MHCA}(\textrm{LN}(\mathbf{D}^{k-1}_{m}), \textrm{LN}(\mathbf{E}^{j}_{m})) + \mathbf{D}^{k-1}_{m}, \\
\mathbf{D}^{k''}_{m}  &= \textrm{MHSA}(\textrm{LN}(\mathbf{D}^{k'}_{m})) + \mathbf{D}^{k'}_{m}, \\
\mathbf{D}^{k}_{m}  &= \textrm{FFN}(\textrm{LN}(\mathbf{D}^{k''}_{m})) + \mathbf{D}^{k''}_{m}, \\
\end{split}
\label{eq_skip_connection}
\end{equation}
where $k \in \{2,...,N_d\}$ is the decoder layer index, $j \in \{q_1, ..., q_{N_c}\}$ is the selected encoder layer index, $N_c$ is the number of skip connections, and $q_i \in \{1, ..., N_s\}$.

\textbf{Reconstruction Loss}. Finally, the decoder output $\mathbf{D}^{N_d}_{m}$ is passed through a linear projection layer to predict the masked information in each modality. We calculate the mean squared error between the ground truth and reconstructed data in the masked positions for each modality and sum them to get the reconstruction loss:
\begin{equation}
\begin{split}
\mathcal{L}_{\textrm{MAE}} &= \mathcal{L}_{\textrm{MAE}}^a + \mathcal{L}_{\textrm{MAE}}^v, \\
\mathcal{L}_{\textrm{MAE}}^a &= \frac{1}{|\mathcal{M}_a|}\sum_{p \in \mathcal{M}_a}|| \mathbf{X}_a(p) - \hat{\mathbf{X}}_a(p)||^2, \\
\mathcal{L}_{\textrm{MAE}}^v &= \frac{1}{|\mathcal{M}_v|}\sum_{p \in \mathcal{M}_v}|| \mathbf{X}_v(p) - \hat{\mathbf{X}}_v(p)||^2, \\
\end{split}
\label{eq_loss_mae}
\end{equation}
where $\mathbf{X}_m$ ($m \in \{a,v\}$) denotes the original input, $\hat{\mathbf{X}}_m$ is the reconstructed data, $\mathcal{M}_m$ denotes the masked positions.
Note that, for the visual modality, we follow \cite{sun2023mae} to reconstruct both spatial appearance and temporal motion information (i.e., frame difference signals).

\subsection{Hierarchical Cross-Modal Contrastive Learning}
\label{sec_method_pretrain_contrastive_learning}
The natural audio-visual correspondences in videos offer a free and useful signal for learning self-supervised representations. Therefore, in addition to masked reconstruction, HiCMAE also utilizes Hierarchical Cross-Modal Contrastive Learning (HCMCL) as a supplement for improved self-supervised audio-visual pre-training (Fig. \ref{fig_overview}). The main benefits of HCMCL are two-fold: 1) HCMCL can narrow the representation gap between audio and video modalities; 2) HCMCL enables better audio-visual information fusion in the subsequent cross-modal encoder. Note that, as shown in Fig. \ref{fig_overview}, different from conventional contrastive learning which is only applied to high-level (i.e., the last layer) features, HCMCL is imposed on multiple intermediate (including both high-level and low-level) features in audio and video encoders to achieve latent representation alignment in a progressive manner. 

To align with hierarchical skip connections in Section \ref{section_decoder}, we use the same $N_c$ selected encoder layers ($i.e., \{q_1, ..., q_{N_c}\}$, $q_i \in \{1, ..., N_s\}$) for HCMCL. 
Since HCMCL is conducted within a batch of samples, for convenience, we add an additional subscript $i$ to the notation of encoder tokens (i.e., $\mathbf{E}^{j}_{i, m}$) to indicate the sample index in this batch. 
Given a batch of $N$ audio tokens $\{\mathbf{E}^{j}_{1, a}, ..., \mathbf{E}^{j}_{N, a}\}$ and $N$ video tokens $\{ \mathbf{E}^{j}_{1, v}, ..., \mathbf{E}^{j}_{N, v} \}$ from $j$th ($j \in \{q_1, ..., q_{N_c}\}$) encoder layer, we first perform global average pooling to obtain the sample-level features, i.e., $\mathbf{e}^{j}_{i,a} = \textrm{AvgPool}(\mathbf{E}^{j}_{i, a})$ and $\mathbf{e}^{j}_{i,v} = \textrm{AvgPool}(\mathbf{E}^{j}_{i, v})$. 
After that, we utilize symmetric InfoNCE loss \cite{oord2018representation} for HCMCL. 
For a batch of audio features $\mathbf{e}^{j}_{a} = \{ \mathbf{e}^{j}_{1, a}, ..., \mathbf{e}^{j}_{N, a} \}$ and video features $\mathbf{e}^{j}_{v} = \{ \mathbf{e}^{j}_{1, v}, ..., \mathbf{e}^{j}_{N, v} \}$, the symmetric InfoNCE loss maximizes the cosine similarity of $N$ paired audio-visual features (i.e., from the same sample) in the batch while minimizing the cosine similarity of features of $N(N-1)$ incorrect pairings (i.e., from different samples):
\begin{equation}
\begin{split}
& \mathcal{L}_{\textrm{InfoNCE}}(\mathbf{e}^{j}_{a}, \mathbf{e}^{j}_{v}) = -\frac{1}{2}[\mathcal{L}(\mathbf{e}^{j}_{a}, \mathbf{e}^{j}_{v}) + \mathcal{L}(\mathbf{e}^{j}_{v}, \mathbf{e}^{j}_{a})], \\
& \mathcal{L}(\mathbf{e}^{j}_{a}, \mathbf{e}^{j}_{v}) = -\frac{1}{N} \sum_{i=1}^{N} \log \frac{\exp{(\textrm{sim}(\mathbf{e}^{j}_{i, a}, \mathbf{e}^{j}_{i, v}) / \tau)}}{\sum_{q=1}^{N} \exp{(\textrm{sim}(\mathbf{e}^{j}_{i, a}, \mathbf{e}^{j}_{q, v}) / \tau)}}, \\
& \mathcal{L}(\mathbf{e}^{j}_{v}, \mathbf{e}^{j}_{a}) = -\frac{1}{N} \sum_{i=1}^{N} \log \frac{\exp{(\textrm{sim}(\mathbf{e}^{j}_{i, v}, \mathbf{e}^{j}_{i, a}) / \tau)}}{\sum_{q=1}^{N} \exp{(\textrm{sim}(\mathbf{e}^{j}_{i, v}, \mathbf{e}^{j}_{q, a}) / \tau)}},
\end{split}
\label{eq_loss_infonce}
\end{equation}
where $\textrm{sim}(\mathbf{x}, \mathbf{y})=\frac{\mathbf{x}^\top \mathbf{y}}{||\mathbf{x}|| \cdot ||\mathbf{y}||}$ is the cosine similarity between $\mathbf{x}$ and $\mathbf{y}$, $\tau$ is the temperature factor. Finally, we sum the InfoNCE loss across $N_c$ selected encoder layers to obtain the overall contrastive loss for HCMCL: 
\begin{equation}
\mathcal{L}_{\textrm{HCMCL}} = \sum_{j=1}^{N_c} \mathcal{L}_{\textrm{InfoNCE}}(\mathbf{e}^{j}_{a}, \mathbf{e}^{j}_{v}). 
\label{eq_loss_hcmcl}
\end{equation}

\subsection{Overall Pre-training Loss}
\label{sec_method_pretrain_loss}
By combining masked audio-visual reconstruction and hierarchical cross-modal contrastive learning, we obtain the overall loss for self-supervised audio-visual pre-training as follows:
\begin{equation}
\mathcal{L} = \mathcal{L}_{\textrm{MAE}} + \lambda \mathcal{L}_{\textrm{HCMCL}},
\label{eq_loss_overall}
\end{equation}
where $\lambda$ is the weight factor for balancing the hierarchical contrastive loss.

During self-supervised pre-training on massive unlabeled AVER data, the audio-visual emotion semantics are implicitly modeled by $\mathcal{L}$ in Eq. (\ref{eq_loss_overall}).
Despite implicit modeling, the visualization analysis of masked audio-visual reconstruction in Section \ref{sec_exp_reconstruction_visualization} shows that the emotion-related information (e.g., smiles in video frames and harmonics in the audio spectrogram) can be well restored by reasoning in limited visible contexts, indicating that HiCMAE can capture audio-visual emotion semantics via large-scale self-supervised pre-training.

\subsection{Hierarchical Feature Fusion for Downstream Fine-tuning}
\label{sec_method_finetuning}
After self-supervised pre-training, we discard the lightweight decoders and only use the encoders in HiCMAE for downstream fine-tuning on limited labeled AVER data. 
To benefit downstream emotion recognition tasks, we utilize both cross-modal and unimodal features from the encoders. The cross-modal features come from the cross-modal fusion encoder. We simply perform global average pooling to output tokens from the last fusion layer. The unimodal features come from modality-specific encoders. To fully exploit features of different levels, we use learnable weights to combine features from different audio/video encoder layers. Thus, the overall feature can be obtained as follows:
\begin{equation}
\begin{split}
\mathbf{e} &= \textrm{Concat}(\mathbf{e}^{N_f}_{a \rightarrow v}, \mathbf{e}^{N_f}_{v \rightarrow a}, \mathbf{e}_{a}, \mathbf{e}_{v}), \\
\mathbf{e}_{a} &= \sum_{j=1}^{N_s} \alpha^{j}_a \mathbf{e}^{j}_{a}, \mathbf{e}_{v} = \sum_{j=1}^{N_s} \alpha^{j}_v \mathbf{e}^{j}_{v}, \\
\end{split}
\label{eq_finetuning}
\end{equation}
where $\mathbf{e}^{N_f}_{a \rightarrow v} = \textrm{AvgPool}(\mathbf{E}^{N_f}_{a \rightarrow v})$, $\mathbf{e}^{N_f}_{v \rightarrow a} = \textrm{AvgPool}(\mathbf{E}^{N_f}_{v \rightarrow a})$, $\mathbf{e}^{j}_{m} = \textrm{AvgPool}(\mathbf{E}^{j}_{m})$ ($m \in \{a,v\}$), $\sum_{j=1}^{N_s} \alpha^{j}_m = 1$.

After obtaining the overall feature  $\mathbf{e}$, we simply use a linear layer to project it to get the final emotion prediction $\hat{\mathbf{y}}$. 
For the classification task, we utilize the classic cross-entropy loss for model fine-tuning:
\begin{equation}
\mathcal{L}_{\textrm{CLS}} = -\sum_{k=1}^{K}{y_k \log{\hat{y}_K}},
\label{eq_loss_cls}
\end{equation}
where $\hat{\mathbf{y}} = [\hat{y}_1, ..., \hat{y}_K] \in \mathbb{R}^{K}$ denotes the prediction, $\mathbf{y} = [y_1, ..., y_K] \in \mathbb{R}^{K}$ is the target, and $K$ is the number of emotion categories.
For the regression task, we compute the mean square error between the target and the prediction: 
\begin{equation}
\mathcal{L}_{\textrm{REG}} = || \mathbf{y} - \hat{\mathbf{y}} ||^2.
\label{eq_loss_reg}
\end{equation}
where $\hat{\mathbf{y}} \in \mathbb{R}^{D}$, $\hat{\mathbf{y}} \in \mathbb{R}^{D}$, and $D$ is the number of emotion dimensions.
During downstream fine-tuning, the audio-visual emotion semantics are explicitly modeled by $\mathcal{L}_{\textrm{CLS}}$ in Eq. (\ref{eq_loss_cls}) or $\mathcal{L}_{\textrm{REG}}$ in Eq. (\ref{eq_loss_reg}).

\section{Experiments}

\subsection{Implementation Details}

We develop three versions of HiCMAE (i.e., base: HiCMAE-B, small: HiCMAE-S, tiny: HiCMAE-T) to meet various needs in real-world applications. Their main difference is the size of hidden units ($C=512$, $C=384$, and $C=256$, respectively) in the encoder. For three models, we use $N_s=10$ layers in modality-specific encoders, $N_f=2$ layers in the cross-modal fusion encoder, and $N_d=4$ layers in the lightweight decoder. We introduce three hierarchical skip connections between the modality-specific encoder and the decoder, specifically between the 4th encoder layer and the 2nd decoder layer, the 7th encoder layer and 3rd decoder layer, as well as between the 10th encoder layer and the 4th decoder layer. The hierarchical cross-modal contrastive learning is also applied to these selected audio-visual encoder layers.

We pre-train HiCMAE on a very large audio-visual dataset VoxCeleb2 \cite{chung2018voxceleb2}. It contains more than one million video clips from over six thousand celebrities. VoxCeleb2 is split into a \textit{development} set and a \textit{test} set. In this paper, we only use its development set which has 1,092,009 video clips for self-supervised pre-training. 
For each video clip, we sample 16 consecutive frames with a temporal stride of 4 frames and follow \cite{sun2023mae} to extract $160\times160$ patch in each frame to obtain the video input with a size of $16\times160\times160\times3$. The corresponding audio waveform (2.56s) is also extracted and converted into a 128-dimensional log Mel filterbank feature sequence using a 25ms Hanning window with a hop length of 10ms \cite{gong2023contrastive}, resulting in the audio spectrogram with a size of $256\times128$.  
The loss weight in Eq. (\ref{eq_loss_overall}) is set to $\lambda=0.0025$. The temperature factor for contrastive learning is fixed to $\tau=0.07$.
We pre-train HiCMAE for 100 epochs using four Nvidia Tesla V100 GPUs with a total batch size of 160 and a base learning rate of $3e-4$. It takes about five days to complete the pre-training. 
For downstream tasks, we fine-tune the pre-trained model for 50 or 100 epochs with a total batch size of 56 and a base learning rate of $1e-3$. During inference, we follow \cite{sun2023mae} to uniformly sample two clips from each video and calculate their mean score as the final prediction. Other hyper-parameters for pre-training and fine-tuning can refer to \cite{tong2022videomae, huang2022masked, gong2023contrastive} for details.

\subsection{Categorical Audio-Visual Emotion Recognition: In-the-Wild Setting}

\subsubsection{Datasets}
The basic dataset information is summarized in Table \ref{tab_dataset}.
\begin{table*}[]
\caption{Basic information of nine AVER datasets used in this paper.}
\label{tab_dataset}
\centering
\resizebox{\linewidth}{!}{
\begin{tabular}{lcccccccccccccc}
\toprule
Dataset & Emotion Type  & Acquisition Condition &  \#Subjects & \#Samples & \#Emotions & Evaluation \\
\midrule
\multirow{2}{*}{MAFW\cite{liu2022mafw}}    & Categorical  &  In-the-wild & N/A  & 9,172   &  11 & 5-fold cross-validation (official) \\
&  Categorical     &  In-the-wild & N/A  & 8,996   &  43   & 5-fold cross-validation (official)  \\
DFEW \cite{jiang2020dfew}       &  Categorical &  In-the-wild    & N/A  &  11,697  &  7 & 5-fold cross-validation (official) \\
MER-MULTI \cite{lian2023mer}    &  Categorical &  In-the-wild    & N/A  &  3,784   &  6 & Official split \\
\multirow{2}{*}{CREMA-D \cite{cao2014crema} }    &  Categorical &  Lab-controlled & 91 &  7,442   &  6 & 5-fold cross-validation (subject-independent) \\
&  Categorical &  Lab-controlled & 91 &  4,896   &  4 & The last fold (subject-independent) \\
MSP-IMPROV  \cite{busso2016msp}            &  Categorical &  Lab-controlled & 12 &  7,798   &  4 & 6-fold cross-validation (session-independent)\\
RAVDESS  \cite{livingstone2018ryerson}     &  Categorical &  Lab-controlled & 24 &  1,440   &  8 & 6-fold cross-validation (subject-independent)\\
IEMOCAP  \cite{busso2008iemocap}           &  Categorical &  Lab-controlled & 10 &  5,531   &  4 & 5-fold cross-validation (session-independent) \\
Werewolf-XL  \cite{zhang2021werewolf}       &  Dimensional &  Lab-controlled & 129 &  14,632   & 3 & 5-fold cross-validation (subject-independent) \\
AVCAffe  \cite{sarkar2022avcaffe}           &  Dimensional &  In-the-wild    & 106 &  58,112   & 2 & Official split \\

\bottomrule
\end{tabular}
}
\end{table*}

\textbf{MAFW} \cite{liu2022mafw} is a large-scale multimodal compound in-the-wild affective dataset.
It consists of 10,045 video clips annotated with 11 common emotions (including seven basic emotions, contempt, anxiety, helplessness, and disappointment). Each video clip is also accompanied by several textual sentences to describe the subject’s affective behaviors. The dataset provides an 11-class single-labeled set (9,172 video clips) and a 43-class compound set (8,996 video clips). For model evaluation, we follow the original paper to adopt a 5-fold cross-validation protocol.

\textbf{DFEW} \cite{jiang2020dfew} comprises 16,372 video clips which are extracted from over 1,500 high-definition movies. This dataset presents several challenging characteristics, such as extreme illumination and occlusion. The video clips are annotated with seven basic emotions (i.e., happy, sad, neutral, anger, surprise, disgust, and fear). To align with previous work \cite{jiang2020dfew, sun2023mae, zhang2023transformer}, we perform 5-fold cross-validation on 11,697 single-labeled clips for evaluation.

\textbf{MER-MULTI} \cite{lian2023mer} provides 3,373 training video clips originating from Chinese TV series and movies. The dataset is annotated with six emotions, including neutral, anger, happiness, sadness, worry, and surprise. We follow the original paper to conduct 5-fold cross-validation on 3,373 video clips for hyperparameter tuning and evaluate the model on a held-out test set with 411 video clips.

\subsubsection{Results on MAFW} 
\label{sec_exp_mafw_sota}
\begin{table*}[]

\caption{Comparison with state-of-the-art methods on MAFW (11-class). SSL: self-supervised learning method or not. AN: anger. DI: disgust. FE: fear. HA: happiness. NE: neutral. SA: sadness. SU: surprise. CO: contempt. AX: anxiety. HL: helplessness. DS: disappointment. UAR: unweighted average recall. WAR: weighted average recall. *: do not use pre-trained models for initialization. Throughout the paper, we highlight the best result in \textbf{bold} and \underline{underline} the second best. 
}
\label{tab_mafw_sota_single}
\centering
\resizebox{\linewidth}{!}{
\begin{tabular}{lccccccccccccccccc}
\toprule
\multirow{2}{*}{Method} & \multirow{2}{*}{SSL} & \multirow{2}{*}{Modality} & \multirow{2}{*}{\tabincell{c}{\#Params\\(M)}}  & \multirow{2}{*}{\tabincell{c}{FLOPs\\(G)}}  
& \multicolumn{11}{c}{Accuracy of Each Emotion (\%)}  & \multicolumn{2}{c}{Metric (\%)} \\ 
\cmidrule(lr){6-16} \cmidrule(lr){17-18} 
&  & &  & & AN    & DI    & FE    & HA    & NE    & SA    & SU    & CO   & AX    & HL   & DS   & UAR  & WAR  \\ 
\midrule

Wav2Vec2.0 \cite{baevski2020wav2vec}  & \checkmark  &  A  &  95  &  18  & \textbf{59.01} &  9.39 & 26.08 & 31.47 & 32.04  & 46.52  & 9.91  & 1.69  & 12.23  & 3.05  & 6.04  & 21.59 & 29.69         \\
HuBERT \cite{hsu2021hubert}           & \checkmark  &  A  &  95  &  18  & 54.97 & \underline{15.49} & \underline{31.20} & 28.64 & 36.88  &\textbf{58.39}  & 12.52 & \underline{2.54}  & 12.55  & 5.34  & \textbf{16.48} & \underline{25.00} & 32.60        \\
WavLM-Plus \cite{chen2022wavlm}       & \checkmark  &  A  &  95  &  18  & 55.62 & \textbf{17.21} & \textbf{40.48} & \textbf{36.65} & 36.53  & \underline{57.44}  & 11.12 & 2.12  & 11.35  & \textbf{9.54}  & \underline{11.54} & \textbf{26.33} & \underline{34.07}        \\

\rowcolor{gray!20}
HiCMAE-T    & \checkmark  &  A  &   8 & 1    & 56.05 & 10.64 & 28.80 & 31.55 & 35.83 & 52.11 & 26.26 & 0.85 & 12.77 &\underline{6.11} & 2.75 & 23.98 & 32.85 \\
\rowcolor{gray!20}
HiCMAE-S    & \checkmark  &  A  &  18 & 2    & \underline{58.29} & 11.11 & 27.68 & 31.07 & \underline{37.15} & 53.34 & \textbf{28.22} & 2.12 & \textbf{16.05} & 3.82 & 4.40 & 24.84 & 34.00 \\
\rowcolor{gray!20}
HiCMAE-B    & \checkmark  &  A  &  32 & 4    & 56.99 & 11.27 & 29.60 & \underline{33.98} & \textbf{38.03} & 53.07 & \underline{27.38} & \textbf{2.97} & \underline{14.96} & 1.53 & 2.20 & 24.72 & \textbf{34.11} \\

\midrule

ResNet-18 \cite{he2016deep}       &  $\times$   &  V  &  11  &  -  & 45.02 & 9.25  & 22.51 & 70.69 & 35.94 & 52.25 & 39.04 & 0.00 & 6.67  & 0.00 & 0.00 & 25.58 & 36.65          \\

ViT \cite{dosovitskiy2020image}   &  $\times$   &  V  &  -  &  -  & 46.03 & 18.18 & 27.49 & 76.89 & 50.70 & 68.19 & 45.13 & 1.27 & 18.93 & 1.53 & 1.65 & 32.36 & 45.04          \\

C3D \cite{tran2015learning}       &  $\times$   &  V  &  78 &  39  & 51.47 & 10.66 & 24.66 & 70.64 & 43.81 & 55.04 & 46.61 & 1.68 & 24.34 & 5.73 & 4.93 & 31.17 & 42.25          \\

ResNet-18+LSTM \cite{liu2022mafw} &  $\times$   &  V  &  -  &  -  & 46.25 & 4.70  & 25.56 & 68.92 & 44.99 & 51.91 & 45.88 & 1.69 & 15.75 & 1.53 & 1.65 & 28.08           & 39.38          \\

ViT+LSTM \cite{liu2022mafw}       &  $\times$   &  V  &  -  &  -  & 42.42 & 14.58 & \textbf{35.69} & 76.25 & 54.48 & 68.87 & 41.01 & 0.00 & 24.40 & 0.00 & 1.65 & 32.67           & 45.56          \\

C3D+LSTM \cite{liu2022mafw}       &  $\times$   &  V  &  -  &  -  & 54.91 & 0.47  & 9.00  & 73.43 & 41.39 & 64.92 & 58.43 & 0.00 & 24.62 & 0.00 & 0.00 & 29.75           & 43.76          \\

Former-DFER \cite{zhao2021former} &  $\times$   &  V  &  18 &  9 & 58.23 & 11.45 & 31.29 & 75.06 & 43.07 & 63.81 & 46.02 & 0.42 & 26.22 & 2.88 & 2.25 & 32.79 & 45.31 \\

T-ESFL \cite{liu2022mafw}         &  $\times$   &  V  &  -  &  -  & 62.70 & 2.51  & 29.90 & \textbf{83.82} & \textbf{61.16} & 67.98 & 48.50 & 0.00 & 9.52  & 0.00 & 0.00 & 33.28 & 48.18     \\
T-MEP \cite{zhang2023transformer} &  $\times$   &  V  &  5  &  6 & 52.91 & 17.41 & 28.01 & 80.79 & 49.42 & 58.73 & 49.54 & 0.00 & 26.18 & 2.25 & 3.56 & 33.53 & 47.53 \\

DFER-CLIP \cite{zhao2023dferclip} &  \checkmark &  V  &  153 &  92  & -   & - & - & - & - & - & - & - & - & - & - & 39.89 & 52.55 \\

SVFAP  \cite{sun2023svfap}        & \checkmark  &  V   &  78 &  44 & 64.60 & 25.20 & \underline{35.68} & \underline{82.77} & 57.12 & 70.41 & 58.58 & \underline{8.05} & 32.42 & 8.40 & 9.89 & 41.19 & 54.28        \\

MAE-DFER \cite{sun2023mae}        & \checkmark    &  V   &  85 &  50 & \textbf{67.77} & \underline{25.35} & 34.88 & 77.13 & 58.26 & 71.09 & 57.46 & \textbf{8.90} & \underline{33.08} & \textbf{11.83} & 12.09 & \underline{41.62} & \underline{54.31}        \\

\rowcolor{gray!20}
HiCMAE-T      & \checkmark  &  V  &  8  & 10    & 61.31 & 22.94 & 30.08 & 78.50 & \underline{58.42} & 70.40 & \underline{61.03} & 2.54 & 31.66 & 7.63 & 12.64  & 39.70 & 52.86 \\
\rowcolor{gray!20}
HiCMAE-S      & \checkmark  &  V  &  18 & 20    & 64.64 & \underline{25.35} & 33.08 & 79.45 & 55.51 & \underline{71.58} & 60.12 & 7.47 & 32.52 & 8.69 & \underline{16.03} & 41.31 & 53.41 \\
\rowcolor{gray!20}
HiCMAE-B      & \checkmark  &  V  &  32 & 32    & \underline{64.91} & \textbf{26.29} & 33.76 & 80.26 & 56.69 & \textbf{72.92} & \textbf{61.40} & \underline{8.05} & \textbf{33.52} & \underline{8.78}  &\textbf{ 16.48} & \textbf{42.10} & \textbf{54.84} \\

\midrule
ResNet-18+LSTM \cite{liu2022mafw}  & $\times$ &  A+V  &  -    &  -   & 54.47 & 11.89 & 7.07 & \underline{82.73} & 54.85 & 55.06 & 39.35 & 0.00 & 15.99 & 0.39 & 0.00  & 29.26  & 42.69 \\
C3D+LSTM \cite{liu2022mafw}               & $\times$ &  A+V  &  -    &  -   & 62.47 & 3.17  & 15.74 & 77.30 & 42.20 & 65.30 & 42.67 & 0.00 & 19.14 & 0.00 & 0.00  & 30.47  & 44.15 \\
AMH  \cite{yoon2020attentive}             & $\times$ &  A+V  &  -    &  -   & 51.73 & 18.68 & 28.13 & 79.14 & 52.55 & 52.26 & 46.29 & 0.26 & 29.62 & 1.74 & 2.39 & 32.98 & 48.83 \\
T-ESFL \cite{liu2022mafw}                 & $\times$ &  A+V  &  -    &  -   & 60.73 & 1.26  & 21.40 & 80.31 & \underline{58.24} & \underline{75.31} & 53.23 & 0.00 & 14.93 & 0.00 & 0.00 & 33.35 & 48.70 \\
T-MEP* \cite{zhang2023transformer}        & $\times$ &  A+V  &  61   & 22   & 54.98 & 22.11 & 32.23 & \textbf{82.79} & 50.90 & 62.50 & 49.93 & 0.87 & 29.27 & \textbf{8.09} & 6.70 & 36.40 & 48.17 \\
T-MEP \cite{zhang2023transformer}         & $\times$ &  A+V  &  61   & 22   & 57.04 & 24.85 & \textbf{36.09} & 78.96 & 50.83 & 61.85 & 51.28 & 1.29 & \textbf{38.47} & 6.46 & 1.70 & 37.17 & 51.15 \\
\rowcolor{gray!20}
HiCMAE-T       & \checkmark &  A+V  &  20   & 14    & 67.72 & 24.73 & 34.56 & 75.81 & 55.63 & 73.74 & 56.45 & 2.97 & 29.69 & 6.87 & \underline{13.74} & 40.17 & 53.41 \\
\rowcolor{gray!20}
HiCMAE-S       & \checkmark &  A+V  &  46   & 28    & \underline{67.94} & \underline{26.13} & \underline{36.00} & 75.00 & 56.51 & 73.33 & \underline{58.41} & \textbf{8.47} & \underline{34.39} & 7.25 & \textbf{14.84} & \underline{41.66} & \underline{54.45} \\
\rowcolor{gray!20}
HiCMAE-B       & \checkmark &  A+V  &  81   & 46    & \textbf{69.24} & \textbf{29.73} & 34.72 & 78.32 & \textbf{59.15} & \textbf{77.69} & \textbf{60.65} & \underline{6.78} & 31.11 & \underline{8.02} & \underline{13.74} & \textbf{42.65} & \textbf{56.17} \\

\midrule
\gray{ResNet18+MDRE \cite{yoon2018multimodal}}  & \gray{$\times$} & \gray{A+V+T} &  \gray{-}    &  \gray{-}   & \gray{45.59}  & \gray{9.35}  & \gray{24.30} & \gray{76.31} & \gray{51.10} & \gray{\textbf{74.87}} & \gray{28.82} & \gray{2.08} & \gray{\underline{30.99}} & \gray{0.00} & \gray{0.00} & \gray{31.22} & \gray{48.33} \\

\gray{AMH \cite{yoon2020attentive}} & \gray{$\times$} & \gray{A+V+T} & \gray{-}    & \gray{-}   & \gray{54.91}  & \gray{\textbf{19.41}} & \gray{30.01} & \gray{82.79} & \gray{51.42} & \gray{60.73} & \gray{51.54} & \gray{0.00} & \gray{28.18} & \gray{0.00} & \gray{0.00} & \gray{34.45} & \gray{49.87} \\
\gray{Rajan et al. \cite{rajan2022cross}}   & \gray{$\times$} & \gray{A+V+T} & \gray{-}    & \gray{-}   & \gray{56.10}  & \gray{9.96}  & \gray{\underline{41.58}} & \gray{\underline{84.13}} & \gray{\textbf{60.39}} & \gray{63.95} & \gray{44.59} & \gray{0.00} & \gray{24.26} & \gray{\textbf{2.69}} & \gray{1.76} & \gray{35.40} & \gray{48.78} \\
\gray{T-ESFL \cite{liu2022mafw}}         & \gray{$\times$} & \gray{A+V+T} & \gray{-}    & \gray{-}   & \gray{\textbf{61.89}}  & \gray{1.10}  & \gray{7.69}  & \gray{\textbf{85.90}} & \gray{-}    & \gray{71.87} & \gray{\textbf{62.17}} & \gray{0.00} & \gray{\textbf{36.00}} & \gray{0.00} & \gray{0.00} & \gray{31.00} & \gray{50.29} \\
\gray{T-MEP* \cite{zhang2023transformer}}         & \gray{$\times$} & \gray{A+V+T} & \gray{111} & \gray{25} & \gray{53.03}  & \gray{\underline{19.32}} & \gray{40.65} & \gray{79.94} & \gray{55.89} & \gray{\underline{74.17}} & \gray{53.48} & \gray{\underline{2.15}} & \gray{26.61} & \gray{1.15} & \gray{\underline{5.10}} & \gray{\underline{37.41}} & \gray{\underline{50.96}} \\
\gray{T-MEP \cite{zhang2023transformer}}          & \gray{$\times$} & \gray{A+V+T} & \gray{111} & \gray{25} & \gray{\underline{56.95}}  & \gray{18.19} & \gray{\textbf{42.89}} & \gray{81.62} & \gray{\underline{60.14}} & \gray{71.60} & \gray{\underline{58.22}} & \gray{\textbf{3.21}} & \gray{30.53} & \gray{\underline{2.27}} & \gray{\textbf{7.51}} & \gray{\textbf{39.37}} & \gray{\textbf{52.85}} \\

\bottomrule
\end{tabular}
}
\end{table*}

We first present the results of 11 single-labeled emotions on MAFW \cite{liu2022mafw} in Table \ref{tab_mafw_sota_single}. Compared with state-of-the-art \textit{audio-visual} methods, we observe that the proposed method outperforms them by large margins across three different model scales. For example, HiCMAE-B surpasses the best-performing supervised method T-MEP \cite{zhang2023transformer} by \textbf{+5.48\%} UAR and \textbf{+5.02\%} WAR, setting a new record of 42.65\% UAR and 56.17\% WAR on this dataset. Besides, even the smallest model HiCMAE-T also achieves higher performance (\textbf{+3.00\%} UAR and \textbf{+2.26\%} WAR) than T-MEP, while having \textbf{3}$\times$ fewer parameters and reducing more than \textbf{36\%} FLOPs. More surprisingly, although incorporating the textual information (i.e., descriptive sentences of affective behaviors) which is beneficial to emotion recognition on this dataset \cite{zhao2023dferclip, foteinopoulou2023emoclip}, T-MEP still lags behind the proposed method by a large margin. It also should be noted that T-MEP employs strong pre-trained models (e.g., DeiT \cite{touvron2021training} and RoBERTa \cite{liu2019roberta}) as unimodal encoders and shows degraded performance without using them. Therefore, these comparison results greatly demonstrate the powerful learning capacity of our method and the superiority of large-scale self-supervised pre-training over traditional supervised learning. 

In addition to the overall performance, we also provide the detailed results of each emotion in Table \ref{tab_mafw_sota_single}. As can be seen, our method achieves superior performance across most emotions, such as anger, disgust, sadness, contempt, and disappointment. It is worth noting that the samples of contempt and disappointment account for only 2.57\% and 1.98\% of the total. Some baseline methods completely fail to recognize these samples due to the imbalanced distribution, while our method improves the previous best result by \textbf{+7.18\%} for contempt and \textbf{+8.14\%} for disappointment.

As for \textit{unimodal} results, our method still shows strong performance when compared with state-of-the-art unimodal baselines. Specifically, when only fine-tuning the \textit{audio} encoder, we achieve competitive or even better results than three cutting-edge large pre-trained speech models (i.e., Wav2Vec2 \cite{baevski2020wav2vec}, HuBERT \cite{hsu2021hubert}, and WavLM-Plus \cite{chen2022wavlm}), while requiring significantly fewer parameters and computational costs (e.g., 8M parameters and 1G FLOPS in HiCMAE-T \textit{versus} 95M parameters and 18G FLOPs in HuBERT).
For the \textit{visual} modality, the proposed method achieves the best trade-off between model performance and model size. For example, HiCMAE-B outperforms the previous state-of-the-art self-supervised method MAE-DFER \cite{sun2023mae} which is also pre-trained on VoxCeleb2 by \textbf{+0.48\%} UAR and \textbf{+0.53\%} WAR, while being \textbf{2.6}$\times$ smaller and using \textbf{36\%} fewer FLOPs.

The results of 43 compound emotions on MAFW are shown in Table \ref{tab_mafw_sota_compound}. This task is quite challenging due to the high difficulty of distinguishing similar compound classes (e.g., `anger\_anxiety', `anger\_disgust', and `anger\_disgust\_anxiety') and extremely imbalanced distribution \cite{liu2022mafw}. For audio-visual modalities, when compared with state-of-the-art T-MEP \cite{zhang2023transformer}, our method achieves competitive or slightly better performance in terms of UAR and WAR. When evaluating macro-averaged F1-score and AUC, our largest model shows significant improvement over the best-performing T-ESFL \cite{liu2022mafw} (i.e., 12.16\% \textit{versus} 8.44\% and 85.30\% \textit{versus} 74.13\%). 
For audio modality, our method achieves consistently moderate improvement over three strong baselines. Finally, for visual modality, our method brings similar performance gains when compared with the previous best results.

\begin{table}[t]
\caption{Comparison with state-of-the-art methods on MAFW (43-class). SSL: self-supervised learning method or not. UAR: unweighted average recall. WAR: weighted average recall. MF1: macro-averaged F1-score. AUC: area under curve. *: do not use pre-trained models for initialization. 
}
\label{tab_mafw_sota_compound}
\centering
\resizebox{\linewidth}{!}{
\begin{tabular}{lcccccccccccccccc}
\toprule
Method &     SSL &     Modality &  UAR   & WAR   & MF1   & AUC   \\ 
\midrule

Wav2Vec2.0  \cite{baevski2020wav2vec} & \checkmark  &   A   &  5.27  & 20.38 & -    & -   \\
HuBERT      \cite{hsu2021hubert}      & \checkmark  &   A   &  5.36  & 20.70 & -    & -   \\
WavLM-Plus \cite{chen2022wavlm}       & \checkmark  &   A   &  5.51  & 21.09 & -    & -   \\

\rowcolor{gray!20}
HiCMAE-T         & \checkmark &   A   &  6.00              & 21.05             & \underline{5.73}   & \underline{66.70}  \\

\rowcolor{gray!20}
HiCMAE-S         & \checkmark &   A   &  \underline{6.03}  & \underline{22.30} & \underline{5.73}   & 67.52   \\
\rowcolor{gray!20}
HiCMAE-B         & \checkmark &   A   &  \textbf{6.16}     & \textbf{22.59}    & \textbf{5.74}      & \textbf{69.11} \\

\midrule
ResNet-18 \cite{he2016deep}         & $\times$   &   V   &  6.18  & 23.83 & 4.89 & 62.92 \\
ViT \cite{dosovitskiy2020image}     & $\times$   &   V   &  8.62  & 31.76 & 7.46 & 74.90 \\    
C3D \cite{tran2015learning}         & $\times$   &   V   &  9.51  & 28.12 & 6.73 & 74.54 \\    
ResNet-18+LSTM \cite{liu2022mafw}   & $\times$   &   V   &  6.93  & 26.60 & 5.56 & 68.86 \\
ViT+LSTM  \cite{liu2022mafw}        & $\times$   &   V   &  8.72  & 32.24 & 7.59 & 75.33 \\
C3D+LSTM  \cite{liu2022mafw}        & $\times$   &   V   &  7.34  & 28.19 & 5.67 & 65.65 \\
T-ESFL    \cite{liu2022mafw}        & $\times$   &   V   &  9.15  & 34.35 & 7.18 & 75.63 \\
Former-DFER \cite{zhao2021former}   & $\times$         &   V   & 10.21  & 32.07 &  -   &  -    \\
T-MEP       \cite{zhang2023transformer}  & $\times$    &   V   &  9.50  & 31.54 &  -   &  -    \\
\rowcolor{gray!20}
HiCMAE-T         & \checkmark &   V   & 10.39  & 34.17 & 8.27   & 81.14  \\
\rowcolor{gray!20}
HiCMAE-S         & \checkmark &   V   & \underline{11.15}  & \underline{35.12} & \textbf{9.61}   & \underline{82.72} \\
\rowcolor{gray!20}
HiCMAE-B         & \checkmark &   V   & \textbf{11.54}  & \textbf{36.41} & \underline{9.54}    & \textbf{83.46}  \\
\midrule
ResNet-18+LSTM \cite{liu2022mafw}            & $\times$  &  A+V  &  7.85  & 31.03 & 5.95 & 71.08 \\ 
C3D+LSTM       \cite{liu2022mafw}            & $\times$  &  A+V  &  7.45  & 29.88 & 5.76 & 68.13 \\
T-ESFL         \cite{liu2022mafw}            & $\times$  &  A+V  &  9.93  & 34.67 & 8.44 & 74.13 \\
T-MEP*         \cite{zhang2023transformer}   & $\times$  &  A+V  & 11.51  & 34.11 &  -   &  -    \\
T-MEP          \cite{zhang2023transformer}   & $\times$  &  A+V  & 13.22  & \underline{36.58} &  -   &  -    \\
\rowcolor{gray!20}
HiCMAE-T        & \checkmark &  A+V  & 12.07  & 34.84 & 10.01   & 83.72 \\
\rowcolor{gray!20}
HiCMAE-S        & \checkmark &  A+V  & \textbf{13.47}  & 36.29 & \underline{11.53}   & \underline{84.62}  \\
\rowcolor{gray!20}
HiCMAE-B        & \checkmark &  A+V  & \underline{13.29}  & \textbf{37.36} & \textbf{12.16}   & \textbf{85.30}  \\

\midrule
\gray{ResNet-18+MDRE \cite{yoon2018multimodal}}  & \gray{$\times$}    & \gray{A+V+T} & \gray{9.02}  & \gray{33.64} & \gray{-}   & \gray{-}    \\
\gray{AMH \cite{yoon2020attentive}}              & \gray{$\times$} & \gray{A+V+T} & \gray{10.24} & \gray{35.35} & \gray{-}   & \gray{-}    \\
\gray{Rajan et al. \cite{rajan2022cross}}      & \gray{$\times$} & \gray{A+V+T} & \gray{11.09} & \gray{35.33} & \gray{-}   & \gray{-}    \\
\gray{T-ESFL \cite{liu2022mafw}}            & \gray{$\times$} & \gray{A+V+T} & \gray{9.68}  & \gray{35.02} & \gray{\underline{\textbf{8.65}}} & \gray{\underline{\textbf{74.35}}} \\
\gray{T-MEP*  \cite{zhang2023transformer}}  & \gray{$\times$}          & \gray{A+V+T} & \gray{\underline{13.25}} & \gray{\underline{37.69}} & \gray{-}   & \gray{-}    \\
\gray{T-MEP  \cite{zhang2023transformer}}   & \gray{$\times$}          & \gray{A+V+T} & \gray{\textbf{15.22}} & \gray{\textbf{39.00}} & \gray{-}   & \gray{-}    \\
\bottomrule
\end{tabular}
}
\end{table}

\subsubsection{Results on DFEW and MER-MULTI}

\begin{table}[t]
\caption{Comparison with state-of-the-art methods on DFEW. SSL: self-supervised learning method or not. UAR: unweighted average recall. WAR: weighted average recall. *: do not use pre-trained models for initialization.
}
\label{tab_dfew_sota}
\centering
\resizebox{\linewidth}{!}{
\begin{tabular}{lccccccc}
\toprule
Method  & SSL & Modality  &  \tabincell{c}{\#Params\\(M)} & \tabincell{c}{FLOPs\\(G)} &  UAR             & WAR \\
\midrule
Wav2Vec2.0 \cite{baevski2020wav2vec}   & \checkmark & A &  95 &  18    & 36.15  & 43.05 \\
HuBERT     \cite{hsu2021hubert}        & \checkmark & A &  95 &  18    & 35.98  & 43.24 \\
WavLM-Plus \cite{chen2022wavlm}        & \checkmark & A &  95 &  18    & \textbf{37.78}  & \textbf{44.64} \\

\rowcolor{gray!20}
HiCMAE-T          & \checkmark & A &  8 &  1     & 34.57  & 42.91 \\
\rowcolor{gray!20}
HiCMAE-S          & \checkmark & A & 18 &  2     & 35.72  & 43.49 \\ 
\rowcolor{gray!20}
HiCMAE-B          & \checkmark & A & 32 &  4     & \underline{36.20}  & \underline{44.50} \\

\midrule

C3D \cite{tran2015learning}             &  $\times$  &  V  &   78   &  39  &  42.74 & 53.54  \\
R(2+1)D-18 \cite{tran2018closer}        &  $\times$  &  V  &   33   &  42  &  42.79 & 53.22 \\
3D ResNet-18 \cite{hara2018can}         &  $\times$  &  V  &   33   &  8  & 46.52 & 58.27    \\ 
EC-STFL \cite{jiang2020dfew}            &  $\times$  &  V  &  -  &  8  & 45.35 & 56.51    \\ 
ResNet-18+LSTM \cite{zhao2021former}    &  $\times$  &  V  &  -  &  8  & 51.32 & 63.85    \\ 
ResNet-18+GRU  \cite{zhao2021former}    &  $\times$  &  V  &  -  &  8  & 51.68 & 64.02    \\ 
Former-DFER \cite{zhao2021former}       &  $\times$  &  V  &  18 &  9  & 53.69 & 65.70   \\ 
CEFLNet \cite{liu2022clip}              &  $\times$  &  V  &  13 & -   & 51.14 & 65.35   \\ 
EST \cite{liu2023expression}            &  $\times$  &  V  &  43 &  -  & 53.43 & 65.85   \\
STT \cite{ma2022spatio}                 &  $\times$  &  V  &  -  &  -  & 54.58 & 66.65   \\ 
NR-DFERNet \cite{li2022nr}              &  $\times$  &  V  &  -  &  6  & 54.21 & 68.19   \\
DPCNet \cite{wang2022dpcnet}            &  $\times$  &  V  &  51 & 10  & 57.11 & 66.32   \\ 
IAL \cite{li2023intensity}              &  $\times$  &  V  &  19  &  10 & 55.71 & 69.24   \\ 
M3DFEL \cite{wang2023rethinking}        &  $\times$  &  V  &  -   &  2  & 56.10 & 69.25       \\
T-MEP  \cite{zhang2023transformer}      &  $\times$  &  V  &  5   &  6  & 54.14 & 65.22 \\
DFER-CLIP \cite{zhao2023dferclip}       &  \checkmark  & V & 153  &  92 & 59.61 & 71.25\\
SVFAP  \cite{sun2023svfap}              & \checkmark & V &  78  &  44 & \underline{62.83} & \underline{74.27} \\ 
MAE-DFER \cite{sun2023mae}              & \checkmark & V &  85  &  50 & \textbf{63.41} & \textbf{74.43} \\ 

\rowcolor{gray!20}
HiCMAE-T          & \checkmark & V &   8 &  10    & 59.14  & 71.24  \\
\rowcolor{gray!20}
HiCMAE-S          & \checkmark & V &  18 &  20   & 61.37  & 72.20  \\
\rowcolor{gray!20}
HiCMAE-B          & \checkmark & V &  32 &  32   & 61.92  & 73.10  \\

\midrule

ResNet-18+LSTM \cite{zhang2023transformer}    &  $\times$& A+V   &  - &  -     & 52.41  & 64.32 \\
C3D+LSTM  \cite{zhang2023transformer}         &  $\times$& A+V   &  - &  -     & 53.77  & 65.17 \\
AMH  \cite{yoon2020attentive}                 &  $\times$& A+V   &  - &  -     & 54.48  & 66.51 \\
T-MEP*  \cite{zhang2023transformer}           &  $\times$& A+V   &  61 &  22    & 55.06  & 66.30 \\
T-MEP \cite{zhang2023transformer}             &  $\times$& A+V   &  61 &  22     & 57.16  & 68.85 \\

\rowcolor{gray!20}
HiCMAE-T          & \checkmark& A+V  & 20 &  14    & 60.13  & 72.43 \\
\rowcolor{gray!20}
HiCMAE-S          & \checkmark& A+V  & 46 &  28    & \underline{63.05}  & \underline{74.33} \\
\rowcolor{gray!20}
HiCMAE-B          & \checkmark& A+V  & 81 &  46    & \textbf{63.76}  & \textbf{75.01} \\

\bottomrule
\end{tabular}
}
\end{table}

\begin{table}[]
\caption{Comparison with state-of-the-art methods on MER-MULTI. SSL: self-supervised learning method or not. UAR: unweighted average recall. WF1: weighted F1-score.}
\label{tab_mer2023_sota}
\centering
\resizebox{\linewidth}{!}{
\begin{tabular}{lccccccc}
\toprule
Method  & SSL & Modality  &  \tabincell{c}{\#Params\\(M)} & \tabincell{c}{FLOPs\\(G)} &  UAR             & WF1 \\
\midrule
eGeMAPS \cite{eyben2015geneva} & $\times$ & A &  - &   -  & -  & 17.28 \\
VGGish \cite{hershey2017cnn}   & $\times$ & A &  - &   -  & -  & 40.76 \\
Wav2Vec2.0 \cite{baevski2020wav2vec}    & \checkmark & A &  95 &  18    & 51.36	& 51.48 \\
HuBERT     \cite{hsu2021hubert}         & \checkmark & A &  95 &  18    & 50.32 & 52.70 \\
WavLM-Plus \cite{chen2022wavlm}         & \checkmark & A &  95 &  18    & \textbf{53.43} & 54.16 \\
HuBERT-CH   \cite{zhang2022wenetspeech} & \checkmark & A &  95 &  18    & -     & \textbf{61.16} \\

\rowcolor{gray!20}
HiCMAE-T          & \checkmark & A &  8 &  1    & 48.35  & 51.33 \\
\rowcolor{gray!20}
HiCMAE-S          & \checkmark & A & 18 &  2    & 51.09  & 54.16 \\
\rowcolor{gray!20}
HiCMAE-B          & \checkmark & A & 32 &  4    & \underline{51.43}  & \underline{55.33} \\

\midrule
ResNet-MSCeleb \cite{he2016deep}   & $\times$ & V & 26 &  -   & -    & 40.32    \\
ResNet-ImageNet \cite{he2016deep}  & $\times$ & V & 26 &  -   & -    & 44.91   \\
SENet-FER2013 \cite{hu2018squeeze} & $\times$ & V & 28 &  -   & -    & 56.69    \\
ResNet-FER2013 \cite{he2016deep}   & $\times$ & V & 26 &  -   & -    & 57.44    \\
MANet-RAFDB \cite{zhao2021learning}& $\times$ & V & 51 &  -   & -    & 56.19   \\

\rowcolor{gray!20}
HiCMAE-T          & \checkmark & V &   8 &  10   & 50.52  & 58.37  \\
\rowcolor{gray!20}
HiCMAE-S          & \checkmark & V &  18 &  20   & \underline{51.53}  & \underline{59.25}  \\
\rowcolor{gray!20}
HiCMAE-B          & \checkmark & V &  32 &  32   & \textbf{52.31}  & \textbf{59.87}  \\

\midrule

\tabincell{c}{ResNet-FER2013+\\HuBERT-CH \cite{lian2023mer}}   & \checkmark & A+V & 121 &  -   & -    & 69.11    \\
\tabincell{c}{MANet-RAFDB+\\HuBERT-CH \cite{lian2023mer}}      & \checkmark & A+V & 146 &  -   & -    & \underline{70.32}   \\

\rowcolor{gray!20}
HiCMAE-T          & \checkmark& A+V  & 20 &  14    & 59.91  & 68.56 \\
\rowcolor{gray!20}
HiCMAE-S          & \checkmark& A+V  & 46 &  28    & \underline{63.18}  & 70.22 \\
\rowcolor{gray!20}
HiCMAE-B          & \checkmark& A+V  & 81 &  46    & \textbf{64.15}  & \textbf{71.33} \\

\bottomrule
\end{tabular}
}
\end{table}

We present the performance comparison with state-of-the-art methods on DFEW \cite{jiang2020dfew} in Table \ref{tab_dfew_sota}. In the \textit{audio-visual} setting, our HiCMAE-S significantly outperforms the previous best-supervised method T-MEP by \textbf{+5.89\%} UAR and \textbf{+5.48\%} WAR, while having \textbf{25\%} fewer parameters and similar computational costs. In \textit{unimodal} settings, although our models underperform state-of-the-art unimodal baselines, they still achieve very competitive performance. It should also be noted that our method has significantly fewer parameters and FLOPs, thus making it more suitable in resource-constrained scenarios. 

The results on a Chinese dataset MER-MULTI \cite{lian2023mer} are shown in Table \ref{tab_mer2023_sota}. Our HiCMAE-B slightly outperforms the previous best method which utilizes MA-Net \cite{zhao2021learning} (supervised on a facial expression dataset) and a powerful HuBERT-CH model (pre-trained on 10k+ hours of Chinese speech data \cite{zhang2022wenetspeech}). The audio-only result of our method has a similar performance with three self-supervised models which are also pre-trained mostly on English speech data, while it is largely inferior to the state-of-the-art HuBERT-CH. We argue that this is mainly due to the large domain gap between pre-training and fine-tuning tasks.
As for video-only results, our method outperforms the previous best-supervised method.

\subsection{Categorical Audio-Visual Emotion Recognition: Lab-Controlled Setting}

\subsubsection{Datasets}
\textbf{CREMA-D} \cite{cao2014crema} is a high-quality audio-visual dataset for studying the multimodal expression and perception of acted emotions. It consists of 7,442 video clips recorded by 91 actors. The video clips are labeled with six emotions, including happiness, sadness, anger, fear, disgust, and neutral state. Since there is no official split, we follow previous studies \cite{sun2023mae, tran2023saaml} to conduct 5-fold cross-validation in a \textit{subject-independent} manner. We also conduct experiments on a subset of four emotions (i.e., happiness, sadness, anger, and neutral state) and only report the result of the last fold in this setup to align with \cite{tran2023saaml}.

\textbf{MSP-IMPROV} \cite{busso2016msp} is an acted audiovisual corpus to explore emotional behaviors during conversational dyadic interactions. The conversations are designed to elicit spontaneous emotions. The dataset contains 8,438 video clips recorded in six sessions from 12 actors. Following \cite{tran2023saaml}, we only use samples of four emotion categories (i.e., anger, happiness, neutral state, and sadness) and conduct 6-fold cross-validation in a \textit{session-independent} manner.
 
\textbf{RAVDESS} \cite{livingstone2018ryerson} is an audio-visual dataset that includes emotional
speech and song. It comprises 2,880 video clips featuring 24 professional actors, each labeled with one of eight emotions (i.e., seven basic emotions and calm). In this paper, we only use the speech part consisting of 1,440 video clips. We adopt a \textit{subject-independent} 6-fold cross-validation protocol for evaluation \cite{su2020msaf, fu2021cross, sun2023mae}.

\textbf{IEMOCAP} \cite{busso2008iemocap} contains about 12 hours of videos from 10 subjects recorded in five sessions. In this paper, we use 5,531 samples of five classes (i.e., anger, neutral state, happiness, excitement, and sadness) and follow the common practice of merging excitement into happiness to formulate a four-emotion classification task \cite{chen2023exploring, tseng2023av}. We conduct 5-fold cross-validation in a \textit{session-independent} manner.

\subsubsection{Results on CREMA-D} 
\begin{table}[]
\caption{Comparison with state-of-the-art methods on CREMA-D (6-class). SSL: self-supervised learning method or not. UAR: unweighted average recall. WAR: weighted average recall.}
\label{tab_cremad_sota_six_cls}
\centering
\resizebox{\linewidth}{!}{
\begin{tabular}{lccccccc}
\toprule
Method  & SSL & Modality  &  \tabincell{c}{\#Params\\(M)} & \tabincell{c}{FLOPs\\(G)} &  UAR             & WAR \\
\midrule

AuxFormer \cite{goncalves2022auxformer}       & $\times$  &  A  &  -  &  -   & -                       & 58.70                    \\
LR+eGeMAPS \cite{keesing2023emotion, eyben2015geneva}    & \checkmark  &  A  &  -  &  -  & 52.70                       & -                    \\
LR+wav2vec \cite{keesing2023emotion, baevski2020wav2vec} & \checkmark  &  A  &  -  &  -  & 66.50                       & -                    \\

Wav2Vec2.0 \cite{baevski2020wav2vec}   & \checkmark & A &  95 &  18    & 72.57  & 72.41 \\
HuBERT     \cite{hsu2021hubert}        & \checkmark & A &  95 &  18    & \underline{72.72}  & \underline{72.57} \\
WavLM-Plus \cite{chen2022wavlm}        & \checkmark & A &  95 &  18    & \textbf{73.34}  & \textbf{73.39} \\

\rowcolor{gray!20}
HiCMAE-T          & \checkmark & A &  8 &  1     & 68.99  & 68.83 \\
\rowcolor{gray!20}
HiCMAE-S          & \checkmark & A & 18 &  2     & 70.84  & 70.70 \\
\rowcolor{gray!20}
HiCMAE-B          & \checkmark & A & 32 &  4     & 71.11  & 71.01 \\

\midrule

AuxFormer \cite{goncalves2022auxformer}       & $\times$  &  V  &  -  &  -   & -                       & 53.10                    \\
VO-LSTM \cite{ghaleb2019multimodal}           & $\times$  &  V  &  -  &  -   & -                       & 66.80                    \\
Goncalves et al. \cite{goncalves2022robust}   & $\times$  &  V  &  -  &  -   & -                       & 62.20                    \\
Lei et al. \cite{lei2023audio}                & $\times$  &  V  &  -  &  -   & 64.68                   & 64.76                    \\

SVFAP  \cite{sun2023svfap}                  & \checkmark & V &  78  &  44  & \underline{77.31}          & \underline{77.37}                    \\ 
MAE-DFER  \cite{sun2023mae}                 & \checkmark & V &  85  &  50  & \textbf{77.33}          & \textbf{77.38}                    \\ 

\rowcolor{gray!20}
HiCMAE-T          & \checkmark & V &   8 &  10   & 74.06  & 73.98  \\
\rowcolor{gray!20}
HiCMAE-S          & \checkmark & V &  18 &  20   & 76.83  & 76.76  \\
\rowcolor{gray!20}
HiCMAE-B          & \checkmark & V &  32 &  32   & 77.25  & 77.21  \\

\midrule

EF-GRU \cite{tran2022pre}                     & $\times$  & A+V &  -  &  -   & -                       & 57.06                    \\
LF-GRU \cite{tran2022pre}                     & $\times$  & A+V &  -  &  -   & -                       & 58.53                    \\
TFN \cite{zadeh2017tensor}                    & $\times$  & A+V &  -  &  -   & -                       & 63.09                    \\
MATER \cite{ghaleb2020multimodal}             & $\times$  & A+V &  -  &  -   & -                       & 67.20                   \\
MulT Base \cite{tran2022pre}                  & \checkmark& A+V &  38  &  -   & -                       & 68.87                    \\ 
MulT Large \cite{tran2022pre}                 & \checkmark& A+V &  89  &  -   & -                       & 70.22                    \\ 
AuxFormer  \cite{goncalves2022auxformer}      & $\times$  & A+V &  -  &  -   & -                       & 71.70                    \\
AV-LSTM \cite{ghaleb2019multimodal}           & $\times$  & A+V &  -  &  -   & -                       & 72.90                    \\  
AV-Gating \cite{ghaleb2019multimodal}         & $\times$  & A+V &  -  &  -   & -                       & 74.00                    \\
Goncalves et al. \cite{goncalves2022robust}   & $\times$  & A+V &  -  &  -   & -                       & 77.30                    \\
Ladder Networks \cite{goncalves2023learning}  & $\times$  & A+V &  -  &  -   & -                       & 80.30                    \\
\tabincell{l}{VQ-MAE-AV+\\Attn. Pooling \cite{sadok2023vector}}   & \checkmark& A+V  & 30 &  -     & -  & 78.40 \\
\tabincell{l}{VQ-MAE-AV+\\Query2Emo \cite{sadok2023vector}}       & \checkmark& A+V  & 30 &  -     & -  & 80.40 \\
\rowcolor{gray!20}
HiCMAE-T          & \checkmark& A+V  & 20 &  14    & 83.84  & 83.74 \\
\rowcolor{gray!20}
HiCMAE-S          & \checkmark& A+V  & 46 &  28    & \underline{84.46}  & \underline{84.38} \\
\rowcolor{gray!20}
HiCMAE-B          & \checkmark& A+V  & 81 &  46    & \textbf{84.91}  & \textbf{84.89} \\

\bottomrule
\end{tabular}
}
\end{table}

\begin{table}[]
\caption{Comparison with state-of-the-art methods on CREMA-D (4-class).}
\label{tab_cremad_sota_four_cls}
\centering
\resizebox{\linewidth}{!}{
\begin{tabular}{lccccccc}
\toprule
Method  & SSL & Modality  &  \tabincell{c}{\#Params\\(M)} & \tabincell{c}{FLOPs\\(G)} &  UAR             & WAR \\
\midrule

Tran et al. \cite{tran2022pre}           & \checkmark & A+V &  -  &  -   & 83.29 & 83.46    \\
AuxFormer  \cite{goncalves2022auxformer} & $\times$   & A+V &  -  &  -   & 91.10 & 91.62    \\
AV-HuBERT \cite{shi2022learning} & \checkmark & A+V & 103 &  -   & -     & 85.47    \\
FAV-HuBERT \cite{tran2023saaml}  & \checkmark & A+V & 103 &  -   & 87.34 & 87.61    \\
TAPT-HuBERT \cite{tran2023saaml} & \checkmark & A+V & 103 &  -   & 92.78 & 92.84    \\
CTAPT-HuBERT \cite{tran2023saaml}& \checkmark & A+V & 103 &  -   & 90.52 & 90.39    \\
AW-HuBERT   \cite{tran2023saaml} & \checkmark & A+V & 103 &  -   & \underline{93.65} & \underline{93.65}    \\

\rowcolor{gray!20}
HiCMAE-T          & \checkmark& A+V  & 20 &  14    & 92.47  & 92.67 \\
\rowcolor{gray!20}
HiCMAE-S          & \checkmark& A+V  & 46 &  28    & 93.34  & 93.48 \\
\rowcolor{gray!20}
HiCMAE-B          & \checkmark& A+V  & 81 &  46    & \textbf{94.00}  & \textbf{94.13} \\

\bottomrule
\end{tabular}
}
\end{table}

\begin{table}[]
\caption{Comparison with state-of-the-art methods on MSP-IMPROV.}
\label{tab_msp_improv_sota}
\centering
\resizebox{\linewidth}{!}{
\begin{tabular}{lccccccc}
\toprule
Method  & SSL & Modality  &  \tabincell{c}{\#Params\\(M)} & \tabincell{c}{FLOPs\\(G)} &  UAR             & WAR \\
\midrule

Tran et al. \cite{tran2022pre}         & \checkmark & A+V &  -  &  -   & 59.41 & 65.29    \\
AuxFormer                              & $\times$   & A+V &  -  &  -   & 62.97 & 70.28    \\
AV-HuBERT   \cite{shi2022learning}     & \checkmark & A+V & 103 &  -   & -     & 65.27    \\
FAV-HuBERT   \cite{tran2023saaml}      & \checkmark & A+V & 103 &  -   & 61.05 & 68.35    \\
TAPT-HuBERT  \cite{tran2023saaml}      & \checkmark & A+V & 103 &  -   & 63.95 & 70.46    \\
CTAPT-HuBERT \cite{tran2023saaml}      & \checkmark & A+V & 103 &  -   & 60.83 & 68.02    \\
AW-HuBERT    \cite{tran2023saaml}      & \checkmark & A+V & 103 &  -   & \underline{65.72} & 71.80    \\

\rowcolor{gray!20}
HiCMAE-T          & \checkmark& A+V  & 20 &  14    & 63.16  & 72.78 \\
\rowcolor{gray!20}
HiCMAE-S          & \checkmark& A+V  & 46 &  28    & 63.90  & \underline{74.35} \\
\rowcolor{gray!20}
HiCMAE-B          & \checkmark& A+V  & 81 &  46    & \textbf{65.78} & \textbf{74.95} \\

\bottomrule
\end{tabular}
}
\end{table}

The performance comparison with state-of-the-art methods on CREMA-D (6-class) \cite{cao2014crema} is presented in Table \ref{tab_cremad_sota_six_cls}. In the \textit{audio-visual} setting, we first find that our method outperforms two self-supervised baselines (MulT Base and Large \cite{tran2022pre}) by substantial margins (\textbf{+13\%} WAR). It is worth noting that they are audio-visual Transformers pre-trained on VoxCeleb2 too. However, they rely on features extracted from other models instead of the raw data as input. Thus, critical information might be lost during feature extraction, leading to significantly inferior results.
VQ-MAE-AV \cite{sadok2023vector} is another strong self-supervised baseline. Similar to us, it is based on masked autoencoder (but requires two-stage pre-training) and also pre-trained on VoxCeleb2. When compared with two versions of VQ-MAE-AV, our method shows large performance improvement over them and can be trained in a single stage. Specifically, the smallest HiCMAE-T surpasses the best VQ-MAE-AV by \textbf{+3.34\%} WAR while using \textbf{33\%} fewer parameters. With the increase in model size, HiCMAE-B pushes the performance gap to even larger (\textbf{+4.49\%} WAR), establishing a new state-of-the-art result on this dataset. These results demonstrate the effectiveness of the three-pronged strategy to promote hierarchical feature learning in HiCMAE. 
Finally, our method also outperforms advanced supervised baselines (e.g., Ladder Networks \cite{goncalves2023learning}).

In the \textit{unimodal} setting, we observe that our method still maintains competitive performance. For example, for visual modality, HiCMAE-B is slightly inferior ($<$0.2\% UAR and WAR) to state-of-the-art MAE-DFER \cite{sun2023mae}, but requires significantly fewer parameters (\textbf{62\%}) and computational costs (\textbf{36\%} FLOPs). For audio modality, although the performance gap between HiCMAE-B and WavLM-Plus is larger (about 2\% UAR and WAR), it is still acceptable as the latter is  \textbf{3$\times$} larger than the former and has \textbf{4.5$\times$} more FLOPs.

In addition to the default six emotions on this dataset, we also conduct experiments on a subset with four emotions. 
The audio-visual results are shown in Table \ref{tab_cremad_sota_four_cls}. 
Among these baselines, most are self-supervised methods and the HuBERT series except for AV-HuBERT \cite{shi2022learning} (i.e., FAV-HuBERT \cite{tran2023saaml}, TAPT-HuBERT \cite{tran2023saaml}, CTAPT-HuBERT \cite{tran2023saaml}, and AW-HuBERT \cite{tran2023saaml}) are pre-trained on VoxCeleb2 with industry-level computation resources (32 Tesla-V100 GPUs) for approximately 10 days. When compared with the best-performing AW-HuBERT, our HiCMAE-B still shows slight improvement (\textbf{+0.35\%} UAR and \textbf{+0.48\%} WAR), while being \textbf{21\%} smaller and training-friendly (we only need 4 Tesla-V100 GPUs to pre-train the model for about 5 days). It should also be noted that AW-HuBERT is a semi-supervised method that requires another labeled dataset and unlabeled samples from VoxCeleb2 for affective adaptation to obtain improved performance. Thus, these results amply demonstrate the superiority of the proposed method.

\subsubsection{Results on MSP-IMPROV, RAVDESS, and IEMOCAP}
The audio-visual results of MSP-IMPROV \cite{busso2016msp} are shown in Table \ref{tab_msp_improv_sota}. 
We observe that, when compared with the state-of-the-art AW-HuBERT, HiCMAE-B achieves similar UAR but shows a large improvement (\textbf{+3.15\%}) in terms of WAR. Two smaller versions of HiCMAE have lower UAR. However, their WAR is still higher than AW-HuBERT.

\begin{table}[]
\caption{Comparison with state-of-the-art methods on RAVDESS. SSL: self-supervised learning method or not. UAR: unweighted average recall. WAR: weighted average recall.}
\label{tab_ravdess_sota}
\centering
\resizebox{\linewidth}{!}{
\begin{tabular}{lccccccc}
\toprule
Method  & SSL & Modality  &  \tabincell{c}{\#Params\\(M)} & \tabincell{c}{FLOPs\\(G)} &  UAR             & WAR \\
\midrule

LR+eGeMAPS \cite{keesing2023emotion, eyben2015geneva}    & \checkmark  &  A  &  -  &  -  & 50.30                       & -                    \\
LR+wav2vec \cite{keesing2023emotion, baevski2020wav2vec} & \checkmark  &  A  &  -  &  -  & 68.80                       & -                    \\

Wav2Vec2.0 \cite{baevski2020wav2vec}   & \checkmark & A &  95 &  18    & 73.44  & 74.38 \\
HuBERT     \cite{hsu2021hubert}        & \checkmark & A &  95 &  18    & \underline{74.15}  & \underline{74.37} \\
WavLM-Plus  \cite{chen2022wavlm}       & \checkmark & A &  95 &  18    & \textbf{75.28}  & \textbf{75.36} \\

\rowcolor{gray!20}
HiCMAE-T          & \checkmark & A &  8 &  1    & 69.92  & 71.53 \\
\rowcolor{gray!20}
HiCMAE-S          & \checkmark & A & 18 &  2    & 70.12  & 72.01 \\
\rowcolor{gray!20}
HiCMAE-B          & \checkmark & A & 32 &  4    & 70.38  & 72.29 \\

\midrule

VO-LSTM \cite{ghaleb2019multimodal}           & $\times$  &  V  &  -  &  -   & -                       & 60.50                    \\
3D ResNeXt-50 \cite{su2020msaf}               & $\times$  &  V  & 26  &  -   & -                       & 62.99                    \\

SVFAP    \cite{sun2023svfap}                 & \checkmark & V &  78  &  44  & \underline{75.15}          & \underline{75.01}                    \\ 
MAE-DFER \cite{sun2023mae}                  & \checkmark & V &  85  &  50  & \textbf{75.91}          & \textbf{75.56}                    \\ 

\rowcolor{gray!20}
HiCMAE-T          & \checkmark & V &   8 &  10   & 62.57  & 62.78  \\
\rowcolor{gray!20}
HiCMAE-S          & \checkmark & V &  18 &  20   & 69.01  & 68.54  \\
\rowcolor{gray!20}
HiCMAE-B          & \checkmark & V &  32 &  32   & 71.35  & 70.97  \\

\midrule

AV-LSTM \cite{ghaleb2019multimodal}           & $\times$  & A+V &  -  &  -   & -                       & 65.80                    \\  
AV-Gating \cite{ghaleb2019multimodal}         & $\times$  & A+V &  -  &  -   & -                       & 67.70                    \\
MCBP \cite{fukui2016multimodal}   & $\times$  & A+V & 51  &  -   & -                       & 71.32                    \\
MMTM \cite{joze2020mmtm}          & $\times$  & A+V & 32  &  -   & -                       & 73.12                   \\
MSAF \cite{su2020msaf}                        & $\times$  & A+V & 26  &  -   & -                       & 74.86                   \\
ERANNs \cite{verbitskiy2022eranns}            & $\times$  & A+V & -   &  -   & -                       & 74.80                   \\
CFN-SR \cite{fu2021cross}                     & $\times$  & A+V & 26  &  -   & -                       & 75.76                   \\

MATER \cite{ghaleb2020multimodal}             & $\times$  & A+V &  -  &  -   & -                       & 76.30                   \\
MulT \cite{tsai2019multimodal}                & $\times$  & A+V &  -  &  -   & -                       & 76.60                   \\
AVT \cite{chumachenko2022self}                & $\times$  & A+V &  -  &  -   & -                       & 79.20                   \\

\tabincell{l}{VQ-MAE-AV+\\Attn. Pooling \cite{sadok2023vector}}   & \checkmark& A+V  & 30 &  -     & -  & 83.20 \\
\tabincell{l}{VQ-MAE-AV+\\Query2Emo \cite{sadok2023vector}}       & \checkmark& A+V  & 30 &  -     & -  & 84.80 \\
\rowcolor{gray!20}
HiCMAE-T          & \checkmark& A+V  & 20 &  14    & 86.26  & 86.11 \\
\rowcolor{gray!20}
HiCMAE-S          & \checkmark& A+V  & 46 &  28    & \underline{86.85}  & \underline{86.67} \\
\rowcolor{gray!20}
HiCMAE-B          & \checkmark& A+V  & 81 &  46    & \textbf{87.96}  & \textbf{87.99} \\

\bottomrule
\end{tabular}
}
\end{table}

\begin{table}[]
\caption{Comparison with state-of-the-art methods on IEMOCAP. SSL: self-supervised learning method or not. UAR: unweighted average recall. WAR: weighted average recall.}
\label{tab_iemocap_sota}
\centering
\resizebox{\linewidth}{!}{
\begin{tabular}{lccccccc}
\toprule
Method  & SSL & Modality  &  \tabincell{c}{\#Params\\(M)} & \tabincell{c}{FLOPs\\(G)} &  UAR             & WAR \\
\midrule
FBANK \cite{tseng2023av}         & $\times$ & A &  -  &  -   & -     & 51.52    \\
AV-HuBERT \cite{shi2022learning} & \checkmark & A &  90 &  -   & -   & 58.54    \\
RepLAI \cite{mittal2022learning} & \checkmark & A &   5 &  -   & -   & 57.53    \\
AVBERT \cite{lee2021parameter}   & \checkmark & A &  10 &  -   & -   & 60.94    \\
MAViL \cite{lee2021parameter}    & \checkmark & A &  86 &  -   & -   & 59.46    \\

Wav2vec 2.0 \cite{baevski2020wav2vec} & \checkmark & A &  95  &  18  & \textbf{69.88}  & \textbf{67.32}    \\
HuBERT \cite{hsu2021hubert}           & \checkmark & A &  95  &  18  & 68.33  & 66.34    \\
WavLM-Plus \cite{chen2022wavlm}       & \checkmark & A &  95  &  18  & \underline{68.64}  & \underline{67.12}    \\

\rowcolor{gray!20}
HiCMAE-T          & \checkmark & A &  8 &  1    & 63.51  & 62.70 \\
\rowcolor{gray!20}
HiCMAE-S          & \checkmark & A & 18 &  2    & 64.67  & 64.06 \\
\rowcolor{gray!20}
HiCMAE-B          & \checkmark & A & 32 &  4    & 65.54  & 65.23 \\

\midrule

HoG \cite{dalal2005histograms}   & $\times$   & V & -   &  -   & -   & 35.83    \\
AV-HuBERT \cite{shi2022learning} & \checkmark & V & 103 &  -   & -   & 26.59    \\
RepLAI \cite{mittal2022learning} & \checkmark & V &  15 &  -   & -   & 40.72    \\
AVBERT \cite{lee2021parameter}   & \checkmark & V &  37 &  -   & -   & 45.80    \\
MAViL \cite{lee2021parameter}    & \checkmark & V &  87 &  -   & -   & 43.03    \\

\rowcolor{gray!20}
HiCMAE-T          & \checkmark & V &   8 &  10   & 46.87 & 49.68  \\
\rowcolor{gray!20}
HiCMAE-S          & \checkmark & V &  18 &  20   & \underline{48.06}  & \underline{50.48}  \\
\rowcolor{gray!20}
HiCMAE-B          & \checkmark & V &  32 &  32   & \textbf{48.11}  & \textbf{50.89}  \\

\midrule

AV-HuBERT \cite{shi2022learning} & \checkmark & A+V & 103 &  -   & -   & 46.45    \\
AVBERT \cite{lee2021parameter}   & \checkmark & A+V &  43 &  -   & -   & 61.87    \\
MAViL \cite{lee2021parameter}    & \checkmark & A+V & 187 &  -   & -   & 54.94    \\
\rowcolor{gray!20}
HiCMAE-T          & \checkmark& A+V  & 20 &  14    & 66.85  & 66.62 \\
\rowcolor{gray!20}
HiCMAE-S          & \checkmark& A+V  & 46 &  28    & \underline{67.46}  & \underline{67.49} \\
\rowcolor{gray!20}
HiCMAE-B          & \checkmark& A+V  & 81 &  46    & \textbf{68.21}  & \textbf{68.36} \\

\bottomrule
\end{tabular}
}
\end{table}

In Table \ref{tab_ravdess_sota}, we compare the proposed method with state-of-the-art methods on RAVDESS \cite{livingstone2018ryerson}. It can be seen that HiCMAE brings large gains over the previous best method. In specific, our HiCMAE-B achieves a gain of \textbf{+3.19\%} WAR over VQ-MAE-AV+Query2Emo \cite{sadok2023vector}. HiCMAE-T still outperforms it by \textbf{+1.31\%} WAR, while using \textbf{33\%} fewer parameters. The unimodal results are less satisfactory on this dataset, probably due to the much smaller model size.


The performance comparison on IEMOCAP \cite{busso2008iemocap} is presented in Table \ref{tab_iemocap_sota}. Most baseline results are from AV-SUPERB \cite{tseng2023av}. We find that HiCMAE achieves substantial improvement over state-of-the-art generic audio-visual representation learner (e.g., MAViL \cite{huang2023mavil} and AVBERT \cite{lee2021parameter}) in the audio-visual setting. For example, our HiCMAE-S surpasses AVBERT by \textbf{+5.62\%} WAR while having a similar model size. HiCMAE-T outperforms MAViL which is also built upon masked autoencoders by \textbf{+11.68\%} WAR while being \textbf{9$\times$} smaller, indicating the importance of reducing domain shift and the benefit of hierarchical feature learning to improve the quality of the learned representations. When evaluating visual-only performance, our method also demonstrates great success. However, the audio-only results are inferior to state-of-the-art large pre-trained speech models, leaving much room for future performance improvement.

\subsection{Dimensional Audio-Visual Emotion Recognition}

\subsubsection{Datasets}
\textbf{Werewolf-XL} \cite{zhang2021werewolf} is an audio-visual database for studying spontaneous emotions during competitive group interactions in Werewolf games. It contains a total of about 15 hours of audio-visual recordings. In this paper, we use 14,632 speakers' samples with dimensional annotations (i.e., arousal, valence, and dominance) and conduct \textit{subject-independent} 5-fold cross-validation for model evaluation. 

\textbf{AVCAffe} \cite{sarkar2022avcaffe} is a large-scale audio-visual affect dataset simulating remote work scenarios. It consists of a total of 108 hours of videos (more than 58,000 video clips) along with self-reported labels for cognitive load and affect (i.e., arousal, and valence). Note that the arousal and valence scores are given on a scale of 1-4 and we follow the original paper to formulate their prediction as a classification task instead of a regression one. This dataset provides an official split (86 subjects for training and 20 subjects for test) for model evaluation. 

\begin{table}[]
\caption{Comparison with state-of-the-art methods on Werewolf-XL. SSL: self-supervised learning method or not. PCC: Pearson correlation coefficient. CCC: concordance correlation coefficient.}
\label{tab_werewolf_sota}
\centering
\resizebox{\linewidth}{!}{
\begin{tabular}{lccccccccc}
\toprule

\multirow{2}{*}{Method} & \multirow{2}{*}{SSL} & \multirow{2}{*}{Modality}
& \multicolumn{2}{c}{Arousal}  & \multicolumn{2}{c}{Valence} & \multicolumn{2}{c}{Dominance} \\ 
\cmidrule(lr){4-5} \cmidrule(lr){6-7} \cmidrule(lr){8-9} 
                       &       &            &    PCC  & CCC  & PCC  & CCC & PCC  & CCC  \\ 

\midrule

eGeMAPS \cite{eyben2015geneva} & $\times$ &  A  & 23.45  & \underline{28.16}  & 8.08 & 6.86  & 31.15 & 34.03    \\
VGGish \cite{hershey2017cnn}   & $\times$ &  A  & 22.88  & \textbf{30.50}  & 5.69 & 3.91  & 29.59 & 31.57    \\

\rowcolor{gray!20}
HiCMAE-T          & \checkmark &  A  & 26.54  & 27.21  & 12.94 & 8.17   & \textbf{37.88} & \textbf{35.83}    \\
\rowcolor{gray!20}
HiCMAE-S          & \checkmark &  A  & \underline{28.40}  & 27.31  & \underline{15.46} & \underline{11.23}  & \underline{37.83} & \underline{35.65}    \\
\rowcolor{gray!20}
HiCMAE-B          & \checkmark &  A  & \textbf{30.04}  & 26.44  & \textbf{17.63} & \textbf{11.93}  & 36.60 & 34.74    \\

\midrule

HOG \cite{dalal2005histograms}  & $\times$ &  V  & 20.82  & 14.43  & 52.54 & 34.56  &
24.76 & 16.90    \\
VGGFace \cite{parkhi2015deep}   & $\times$ &  V  & 7.24  & 4.61  & 62.96 & 60.38  &
14.30 & 8.20    \\

SVFAP   \cite{sun2023svfap}     & \checkmark &  V   & \underline{23.51}  & \textbf{18.96}  & \textbf{67.11} & \textbf{64.27}  & \underline{34.61} & \textbf{29.69}    \\

\rowcolor{gray!20}
HiCMAE-T          & \checkmark &  V   & 22.45  & 17.73  & 66.55 & 62.40  & 33.57 & 28.25    \\
\rowcolor{gray!20}
HiCMAE-S          & \checkmark &  V   & 23.11  & 18.15  & \underline{67.05} & 62.81  & 34.00 & 28.95    \\
\rowcolor{gray!20}
HiCMAE-B          & \checkmark &  V   & \textbf{24.04}  & \underline{18.56}  & 67.03 & \underline{63.30}  & \textbf{34.91} & \underline{29.52}    \\

\midrule

Zhang et al. \cite{zhang2021werewolf}   & $\times$ &  A+V  & 16.41  & 27.70  & 63.14 & 62.34  &
35.40 & 38.40    \\

\rowcolor{gray!20}
HiCMAE-T          & \checkmark &  A+V & 30.47  & 28.98  & 68.50 & 63.23  & \textbf{42.37} & \textbf{39.86}  \\
\rowcolor{gray!20}
HiCMAE-S          & \checkmark &  A+V & \underline{31.08}  & \underline{30.17}  & \underline{68.92} & \underline{64.41}  & \underline{41.38} & \underline{38.90}  \\ 
\rowcolor{gray!20}
HiCMAE-B          & \checkmark &  A+V & \textbf{33.74}  & \textbf{31.85}  & \textbf{69.23} & \textbf{64.81}  & 40.66 & 37.54  \\

\bottomrule
\end{tabular}
}
\end{table}

\begin{table}[]
\caption{Comparison with state-of-the-art methods on AVCAffe. SSL: self-supervised learning method or not. The evaluation metric for arousal and valence is weighted F1-score.}
\label{tab_avcaffe_sota}
\centering
\resizebox{\linewidth}{!}{
\begin{tabular}{lccccccccccc}
\toprule

Method   & SSL &   Modality    &  \tabincell{c}{\#Params\\(M)} & \tabincell{c}{FLOPs\\(G)}   & Arousal             & Valence            \\ \midrule

\tabincell{l}{VGG-16+\\MC3-18 \cite{sarkar2022avcaffe}}          &  $\times$  &   A+V  &  47  &  -   & 38.90  & 41.70  \\
\tabincell{l}{VGG-16+\\3D ResNet-18 \cite{sarkar2022avcaffe}}    &  $\times$  &   A+V  &  69  &  -   & 37.30  & 39.40  \\
\tabincell{l}{VGG-16+\\R(2+1)D-18 \cite{sarkar2022avcaffe}}      &  $\times$  &   A+V  &  67  &  -   & 40.50  & 39.50  \\
\tabincell{l}{ResNet-18+\\MC3-18 \cite{sarkar2022avcaffe}}       &  $\times$  &   A+V  &  44  &  -   & 36.00  & 39.20  \\
\tabincell{l}{ResNet-18+\\3D ResNet-18 \cite{sarkar2022avcaffe}} &  $\times$  &   A+V  &  66  &  -   & 35.10  & 39.10  \\
\tabincell{l}{ResNet-18+\\R(2+1)D-18 \cite{sarkar2022avcaffe}}   &  $\times$  &   A+V  &  64  &  -   & 39.50  & 37.70  \\

\rowcolor{gray!20}
HiCMAE-T          & \checkmark& A+V  & 20 &  14    & 39.64  & 36.74 \\
\rowcolor{gray!20}
HiCMAE-S          & \checkmark& A+V  & 46 &  28    & \underline{42.13}  & \underline{42.65} \\
\rowcolor{gray!20}
HiCMAE-B          & \checkmark& A+V  & 81 &  46    & \textbf{43.18}  & \textbf{44.20} \\

\bottomrule
\end{tabular}
}
\end{table}

\subsubsection{Results} 

In Table \ref{tab_werewolf_sota}, we compare HiCMAE with state-of-the-art methods on Werewolf-XL \cite{zhang2021werewolf}.  It can be seen that our method outperforms baselines by large margins in terms of Pearson correlation coefficient (PCC) and concordance correlation coefficient (CCC). In the audio-visual setting, for example, HiCAME-B surpasses the previous best by \textbf{+17\%} PCC and \textbf{+4\%} CCC in arousal and \textbf{+6\%} PCC and \textbf{+4\%} CCC in valence. In the unimodal setting, HiCMAE also achieves competitive or even better results. Specifically, for the video modality, HiCMAE-B has similar performance with the state-of-the-art SVFAP while having significantly fewer parameters (32M \textit{versus} 78M) and FLOPs (32G \textit{versus} 44G). For the audio modality, HiCMAE brings large performance gains over the best-performing baselines on two evaluation metrics in three emotion dimensions except for CCC in arousal.


The audio-visual performance comparison on AVCAffe \cite{sarkar2022avcaffe} is presented in Table \ref{tab_avcaffe_sota}. We find that HiCMAE-B outperforms the previous best results by \textbf{+2.68\%} weighted F1-score in arousal and \textbf{+3.50\%} weighted F1-score in valence. As for the smaller model HiCMAE-S, although we see some performance drop, it still beats the baselines moderately in two dimensions while having similar parameters with them. 
Finally, HiCMAE-T typically lags behind the previous supervised methods, probably due to its too-small model capacity.

\subsection{Ablation Studies}
In this section, we conduct in-depth ablation studies to investigate several key design factors in HiCMAE. By default, we present the results of HiCMAE-B on two representative datasets, i.e., MAFW (11-class) and CREMA-D (6-class).

\begin{table}[]
\caption{Ablation study on three-pronged strategy for hierarchical feature learning. HSP: hierarchical skip connections. HCMCL: hierarchical cross-modal contrastive learning. HFF: hierarchical feature fusion.}
\label{tab_ablation_main_module}
\centering
\scalebox{0.8}{
\begin{tabular}{cccccccc}
\toprule

\multirow{2}{*}{HSP} &  \multirow{2}{*}{HCMCL} &  \multirow{2}{*}{HFF}  & \multicolumn{2}{c}{MAFW}  & \multicolumn{2}{c}{CREMA-D} \\
\cmidrule(lr){4-5} \cmidrule(lr){6-7} 
& & & UAR    & WAR  & UAR    & WAR  \\
\midrule
$\times$     & $\times$     & $\times$       &  41.06   & 54.79   & 83.76  & 83.64     \\
\checkmark   & $\times$     & $\times$       &  41.82   & 55.45   & 84.30  & 84.26     \\
\checkmark   & \checkmark   & $\times$       &  42.48   & 55.83   & 84.73  & 84.68     \\
\rowcolor{gray!20}
\checkmark   & \checkmark   & \checkmark     &  \textbf{42.65}   & \textbf{56.17}   & \textbf{84.91}  & \textbf{84.89}  \\
\bottomrule
\end{tabular}
}
\end{table}

\subsubsection{Three-pronged Strategy for Hierarchical Feature Learning}
We first investigate the effect of the three-pronged strategy in HiCMAE to promote hierarchical feature learning, including hierarchical skip connections between the encoder and decoder, hierarchical cross-modal contrastive learning, and hierarchical feature fusion for downstream fine-tuning. To this end, we sequentially remove one of them and evaluate the corresponding variant in downstream tasks. The ablation results are shown in Table \ref{tab_ablation_main_module}. 
We observe that the removal of any module will lead to degraded performance and the model achieves the worst performance when all three modules are removed (i.e., the vanilla audio-visual masked autoencoder), which verifies their effectiveness in fostering hierarchical representation learning and improving the overall equality of the learned audio-visual representations in downstream tasks. We also notice that hierarchical skip connections between the encoder and decoder contribute the most among the three modules, followed by hierarchical cross-modal contrastive learning, and finally hierarchical feature fusion for downstream fine-tuning.

\begin{table}[]
\caption{Ablation study on loss weight for self-supervised pre-training.}
\label{tab_ablation_loss_weights}
\centering
\scalebox{0.8}{
\begin{tabular}{cccccccc}
\toprule

\multirow{2}{*}{Loss weight $\lambda$}  & \multicolumn{2}{c}{MAFW}  & \multicolumn{2}{c}{CREMA-D} \\
\cmidrule(lr){2-3} \cmidrule(lr){4-5} 
 & UAR    & WAR  & UAR    & WAR  \\
\midrule
0            &  41.95   & 55.62   & 84.38  & 84.34     \\
0.001        &  42.36   & \textbf{56.37}   & 84.64  & 84.57     \\
\rowcolor{gray!20}
0.0025       &  \textbf{42.65} & 56.17  & \textbf{84.91}  & \textbf{84.89}  \\
0.005        &  42.42   & 55.90   & 84.82  & 84.84     \\
0.01         &  41.94   & 54.98   & 84.70  & 84.63    \\
0.1          &  35.34   & 47.87   & 79.44  & 79.43     \\
\bottomrule
\end{tabular}
}
\end{table}

\subsubsection{Loss Weight}
We then explore the role of contrastive loss weight $\lambda$ in Eq. (\ref{eq_loss_overall}). 
As presented in Table \ref{tab_ablation_loss_weights}, we have the following observations: 1) when $\lambda=0$, the model achieves sub-optimal performance on both datasets, which indicates the necessity of hierarchical cross-modal contrastive learning during self-supervised pre-training. 2) when $\lambda=0.1$, the model achieves significantly worse performance. This implies that masked reconstruction loss is more essential than hierarchical cross-modal contrastive loss in HiCMAE. Large $\lambda$ will overemphasize the latter too much and undermine the former, thus leading to a significant performance decline. 3) HiCMAE works reasonably well when $\lambda \in [0.001, 0.01]$ and it generally achieves the best performance when $\lambda=0.0025$.

\begin{table}[]
\caption{Ablation study on layers in different encoders. $N_s$: number of layers in modality-specific encoders. $N_f$: number of layers in cross-modal fusion encoder.}
\label{tab_ablation_layers_in_unimodal_cross_modal_encoder}
\centering
\scalebox{0.8}{
\begin{tabular}{cccccccc}
\toprule

\multirow{2}{*}{$N_s$} & \multirow{2}{*}{$N_f$}  & \multicolumn{2}{c}{MAFW}  & \multicolumn{2}{c}{CREMA-D} \\
\cmidrule(lr){3-4} \cmidrule(lr){5-6} 
& & UAR    & WAR  & UAR    & WAR  \\
\midrule
12      &    0   &  41.76   & 55.49   & 84.06  & 84.04     \\
11      &    1   &  42.42   & 55.90   & 84.56  & 84.52     \\
\rowcolor{gray!20}
10      &    2   &  \textbf{42.65} & \textbf{56.17}  & \textbf{84.91}  & \textbf{84.89}  \\
8       &    4   &  42.20   & 55.86   & 84.25  & 84.21     \\
\bottomrule
\end{tabular}
}
\end{table}

\subsubsection{Layers in Modality-Specific and Cross-Modal Fusion Encoder}
Next, we ablate the choice of the number of layers in modality-specific encoders ($N_s$) and the cross-modal fusion encoder ($N_f$) by keeping their sum fixed. The ablation results are shown in Table \ref{tab_ablation_layers_in_unimodal_cross_modal_encoder}. 
We first find that the model achieves the worst performance when the cross-modal fusion encoder is removed (i.e., $N_f=0$). This observation demonstrates the crucial role of the cross-modal fusion encoder in integrating heterogeneous audio-visual information. Besides, only one fusion layer is also beneficial to improve the result. Finally, maintaining enough layers in modality-specific encoders is also necessary since too small $N_s$ will also hurt model performance.

\subsubsection{Information Flow in Cross-modal Fusion Encoder}

\begin{figure}[t]
	\centering
    \includegraphics[width=1.0\linewidth]{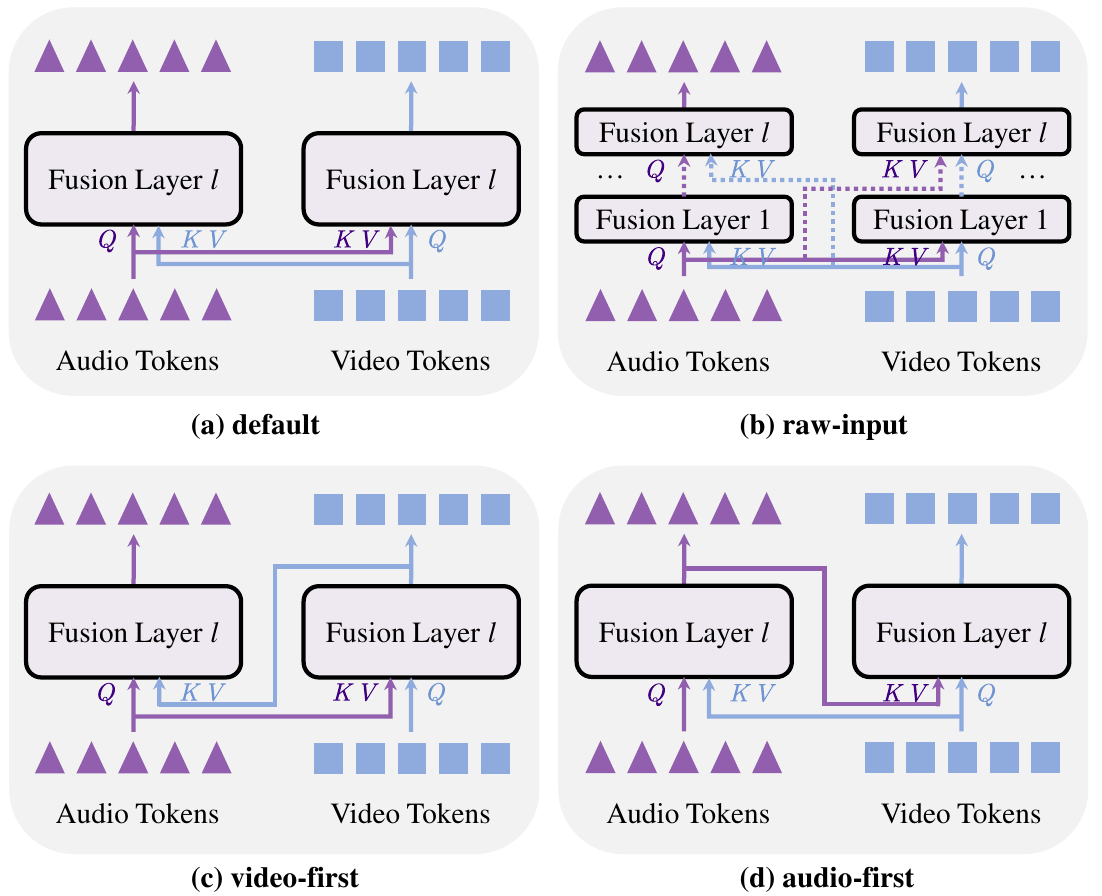}
    \caption{Different types of information flow in cross-modal fusion encoder.}
	\label{fig_cross_modal_information_flow}
\end{figure}

\begin{table}[]
\caption{Ablation study on information flow in cross-modal fusion encoder.}
\label{tab_ablation_information_flow}
\centering
\scalebox{0.8}{
\begin{tabular}{cccccccc}
\toprule

\multirow{2}{*}{Type of information flow}  & \multicolumn{2}{c}{MAFW}  & \multicolumn{2}{c}{CREMA-D} \\
\cmidrule(lr){2-3} \cmidrule(lr){4-5} 
 & UAR    & WAR  & UAR    & WAR  \\
\midrule
\rowcolor{gray!20}
default      &  42.65   & \textbf{56.17}  & \textbf{84.91}  & \textbf{84.89}  \\
raw-input    &  42.30   & 55.86           & 84.47  & 84.43     \\
video-first  &  \textbf{42.74} & 56.06    & 84.81  & 84.78    \\
audio-first  &  42.44   & 56.00           & 84.60  & 84.57     \\
\bottomrule
\end{tabular}
}
\end{table}

\begin{figure}[t]
	\centering
    \includegraphics[width=1.0\linewidth]{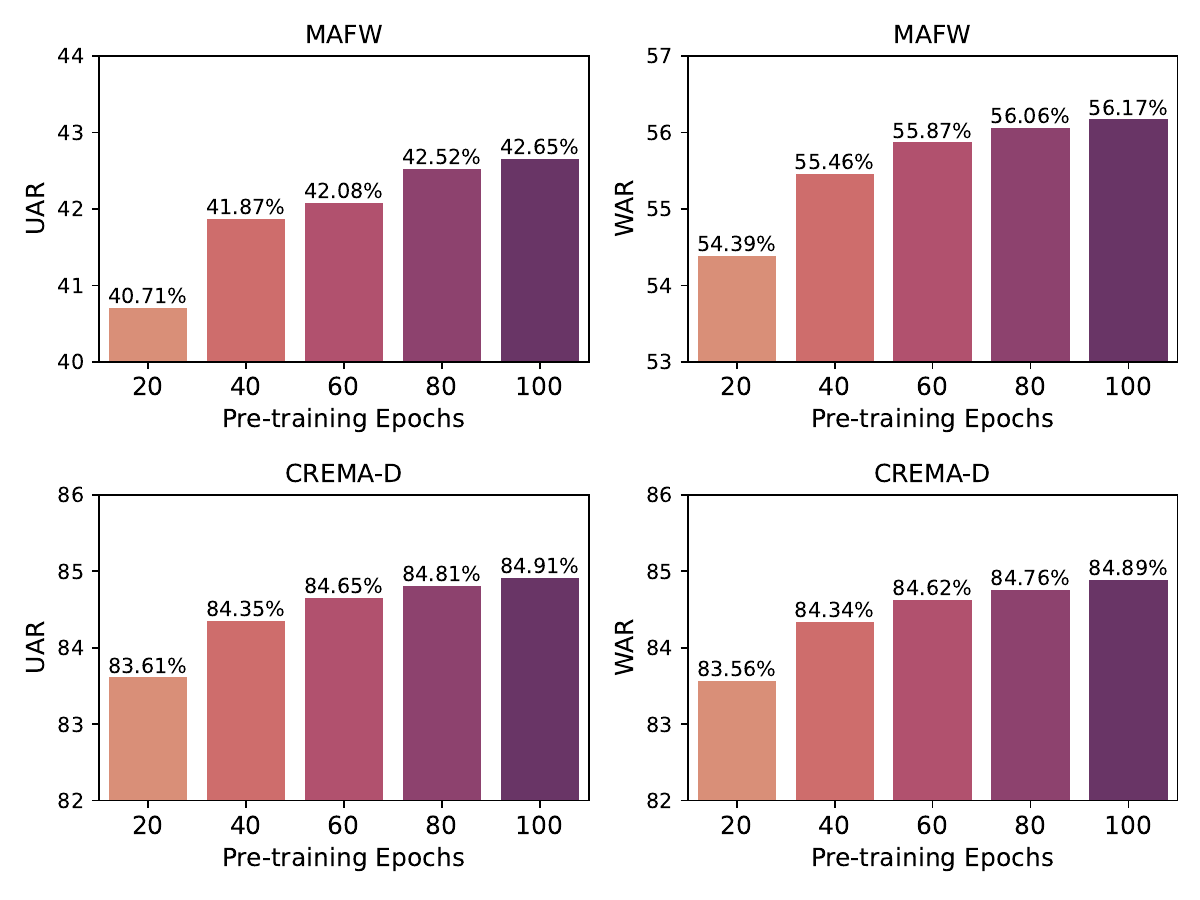}
    \caption{Ablation study on pre-training epochs.}
	\label{fig_ablation_pretraining_epochs}
\end{figure}

\begin{figure}[t]
	\centering
    \includegraphics[width=1.0\linewidth]{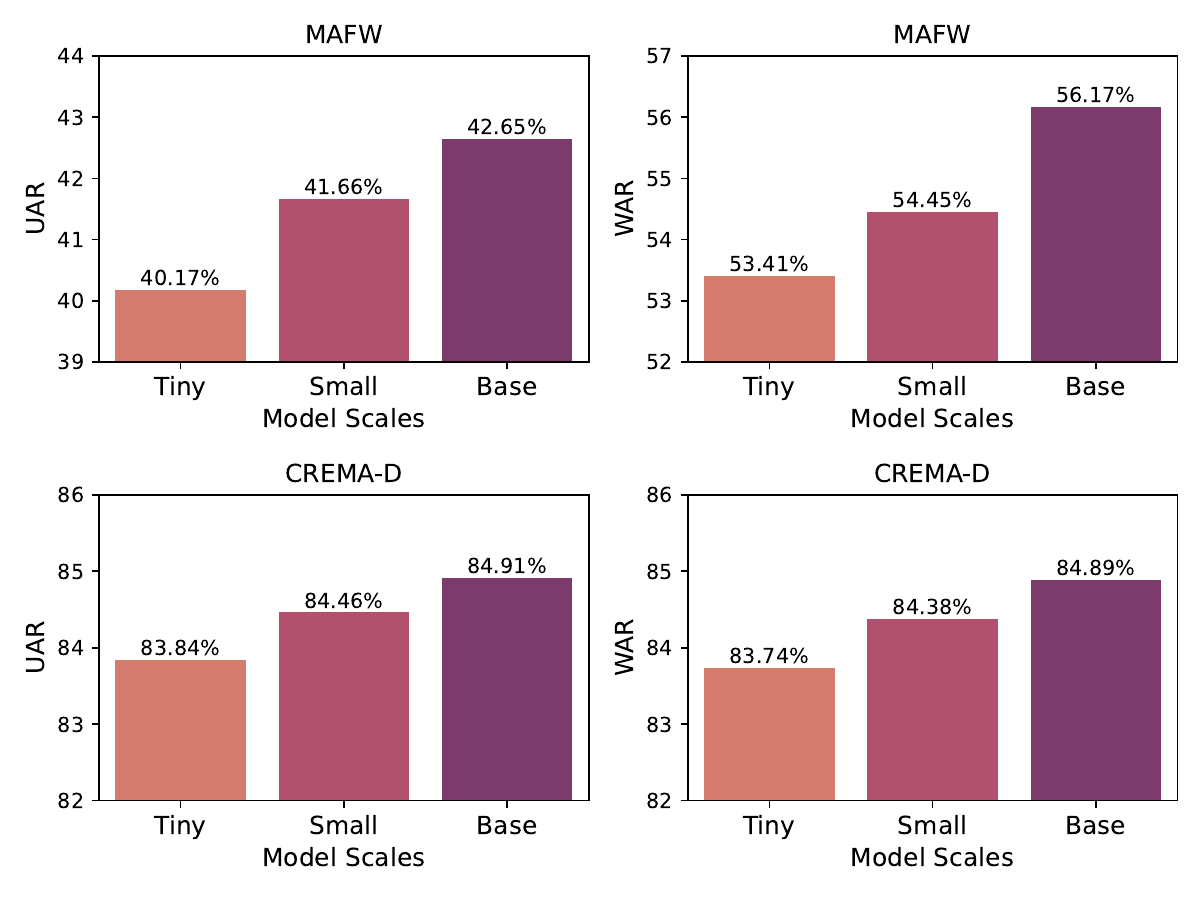}
    \caption{Ablation study on model scales.}
	\label{fig_ablation_model_scales}
\end{figure}

\begin{figure*}[]
	\centering
    \includegraphics[width=1.0\linewidth]{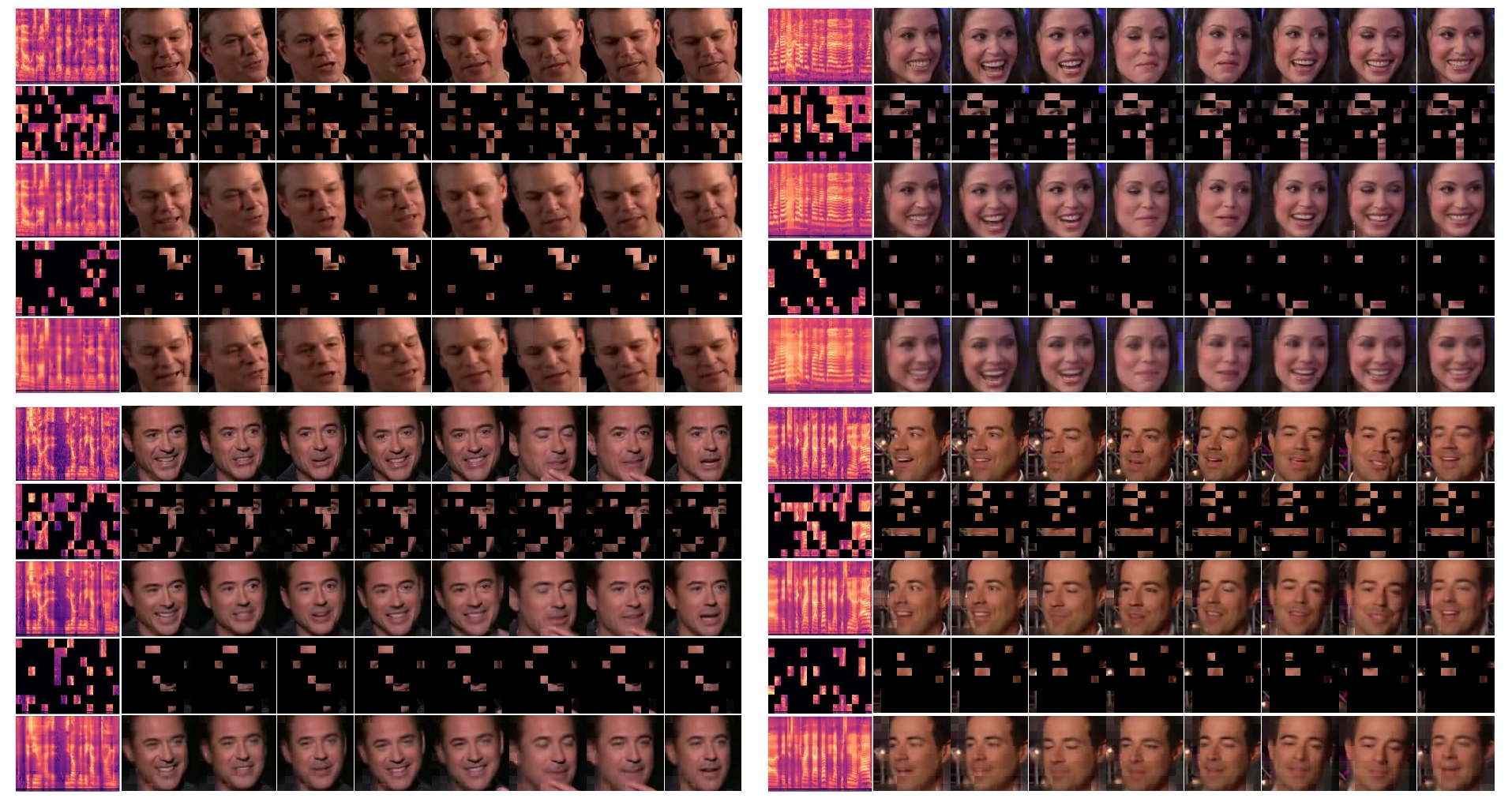}
    \caption{Reconstruction visualization of four unseen celebrities from the test set of VoxCeleb2. For each subject, we sequentially show the original audio spectrogram and video frames, masked inputs in medium level (audio: 60\%, video: 75\%) along with the reconstructed data, and highly masked inputs (audio: 80\%, video: 90\%) along with the reconstructed data. Zoom in to see reconstruction details.}
	\label{fig_masked_reconstruction}
\end{figure*}

\begin{figure*}[]
	\centering
    \includegraphics[width=1.0\linewidth]{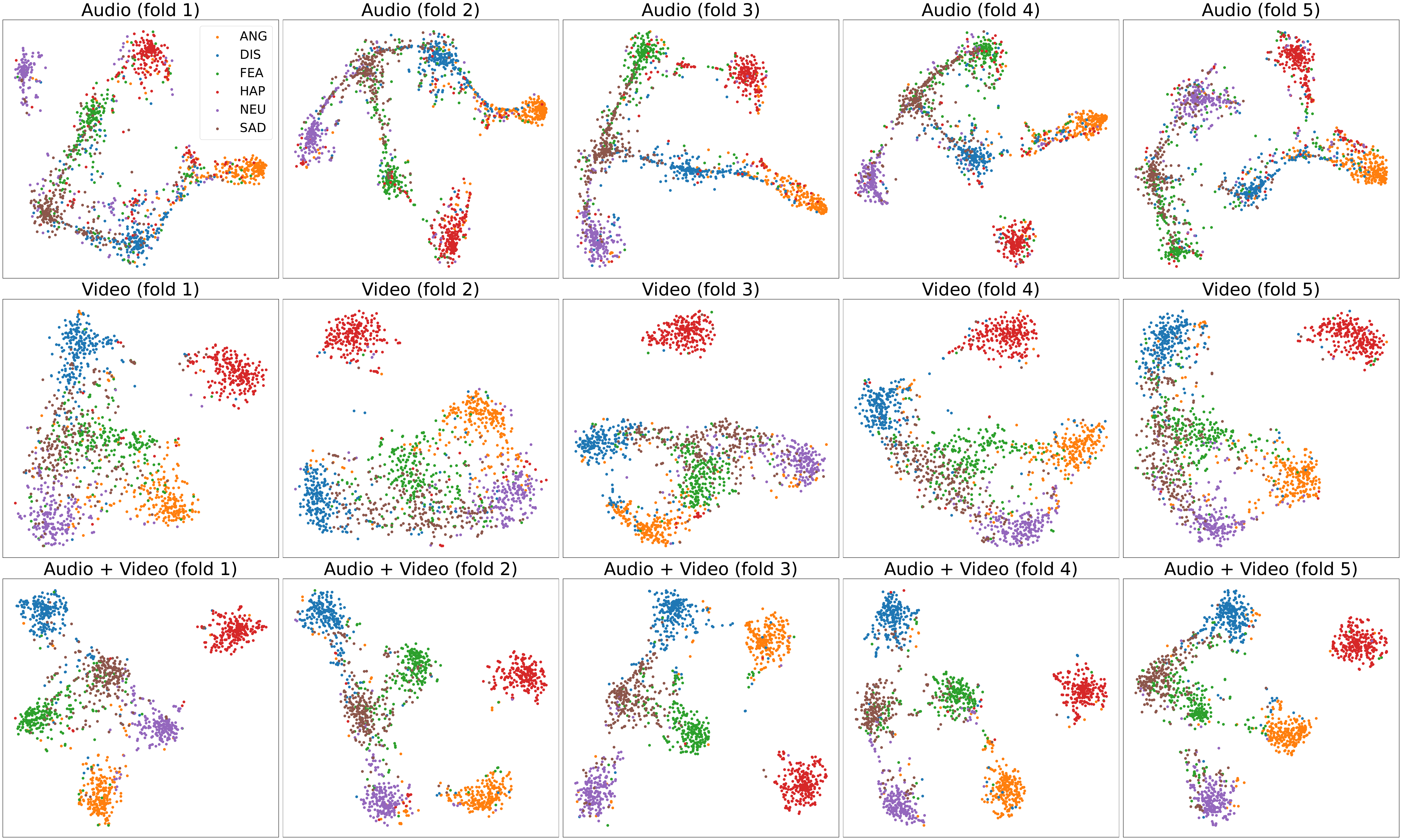}
    \caption{Unimodal and multimodal embedding space visualization on CREMA-D (6-class).}
	\label{fig_embedding_space}
\end{figure*}

We then investigate the effect of different types of information flow in the cross-modal fusion encoder. We develop three variants of the default information flow in Eq. (\ref{eq_fusion_a2v}-\ref{eq_fusion_v2a}) and show their differences in Fig. \ref{fig_cross_modal_information_flow}. Specifically, for the \textit{raw-input} variant, tokens of one modality in each fusion layer always attend to the raw input tokens of the other modality \cite{tsai2019multimodal}, instead of updated tokens from the last layer. For the \textit{video-first} variant, video tokens first update themselves via audio information from the last fusion layer and then audio tokens attend to the updated video tokens. The \textit{audio-first} variant is just the reverse of the video-first variant. The ablation results are presented in Table \ref{tab_ablation_information_flow}. We observe that the model performance is not sensitive to different types of information flow in the cross-modal fusion encoder. Besides, in general, the default information flow works best, followed by the video-first and audio-first variants, and finally the raw-input variant.



\subsubsection{Pre-Training Epochs}
In this part, we explore the effect of pre-training epochs on downstream fine-tuning. The results are shown in Fig. \ref{fig_ablation_pretraining_epochs}. As can be seen, we find that longer pre-training epochs lead to improved fine-tuning performance, which is consistent with our expectations. Besides, the model performance begins to saturate around 80 epochs. Due to limited computation resources and expensive time costs, we stop pre-training at 100 epochs. Nevertheless, we believe that continual pre-training will bring further performance gains and we encourage other researchers who have access to industry-level computation resources to conduct follow-up studies.


\begin{figure*}[]
	\centering
    \subfloat[Total\label{fig_conf_mat_mafw}]{
        \includegraphics[scale=0.35]{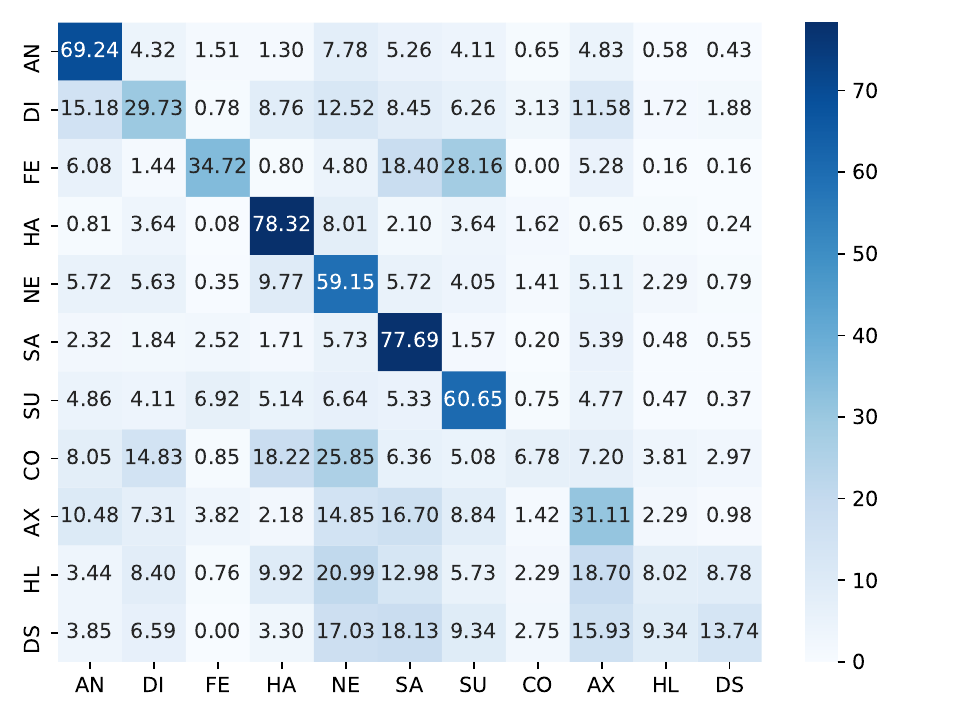}}
	\subfloat[Fold 1\label{fig_conf_mat_mafw_fold_1}]{
		\includegraphics[scale=0.35]{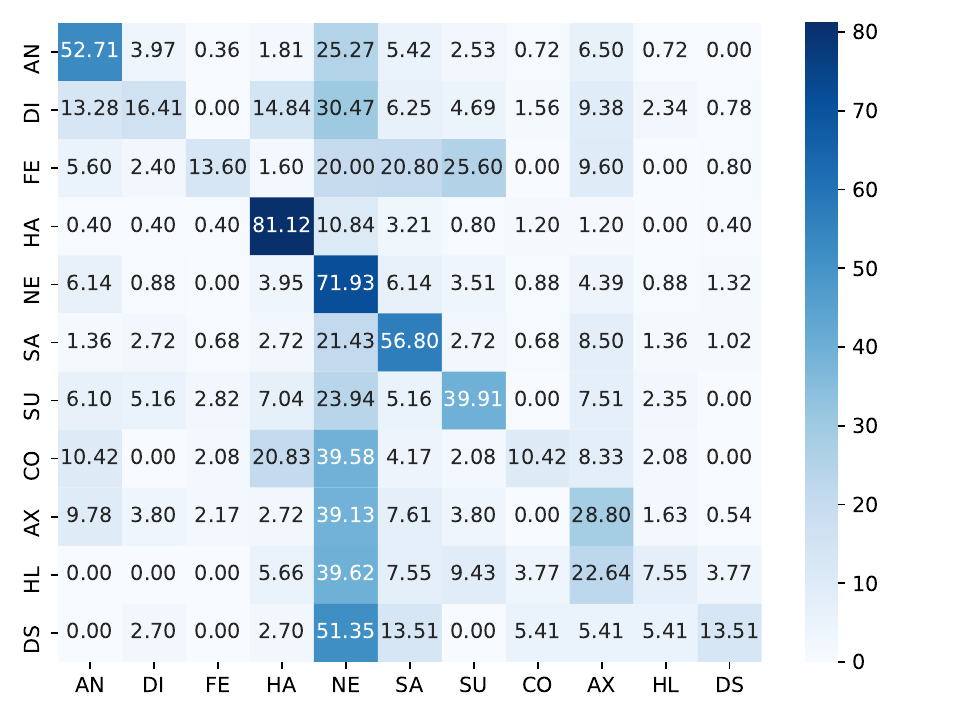}}
    \subfloat[Fold 2\label{fig_conf_mat_mafw_fold_2}]{
        \includegraphics[scale=0.35]{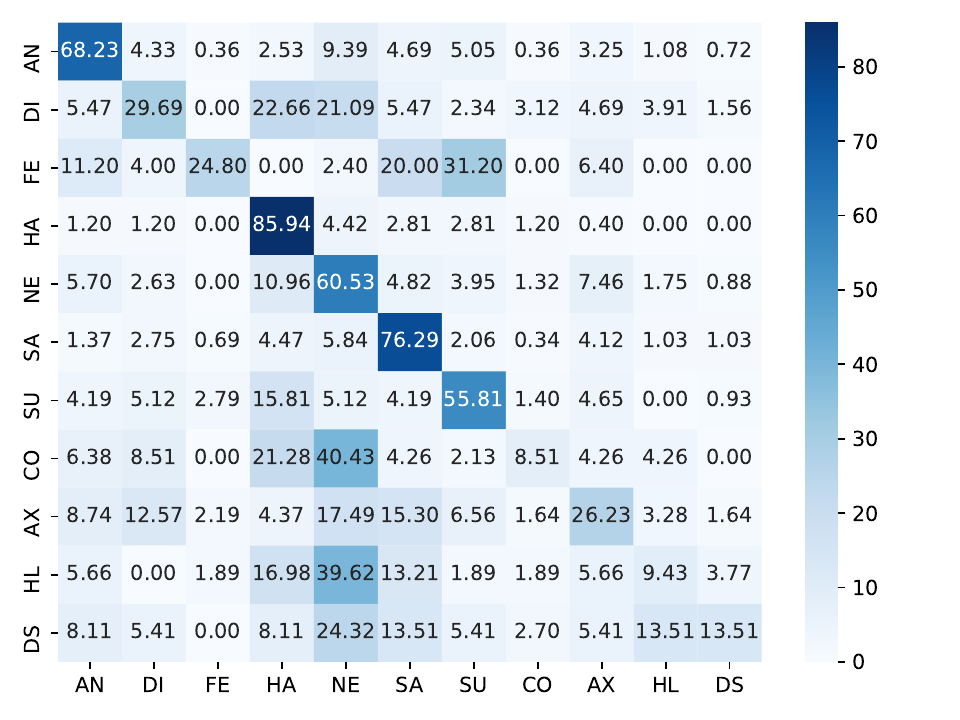}} \\
	\subfloat[Fold 3\label{fig_conf_mat_mafw_fold_3}]{
		\includegraphics[scale=0.35]{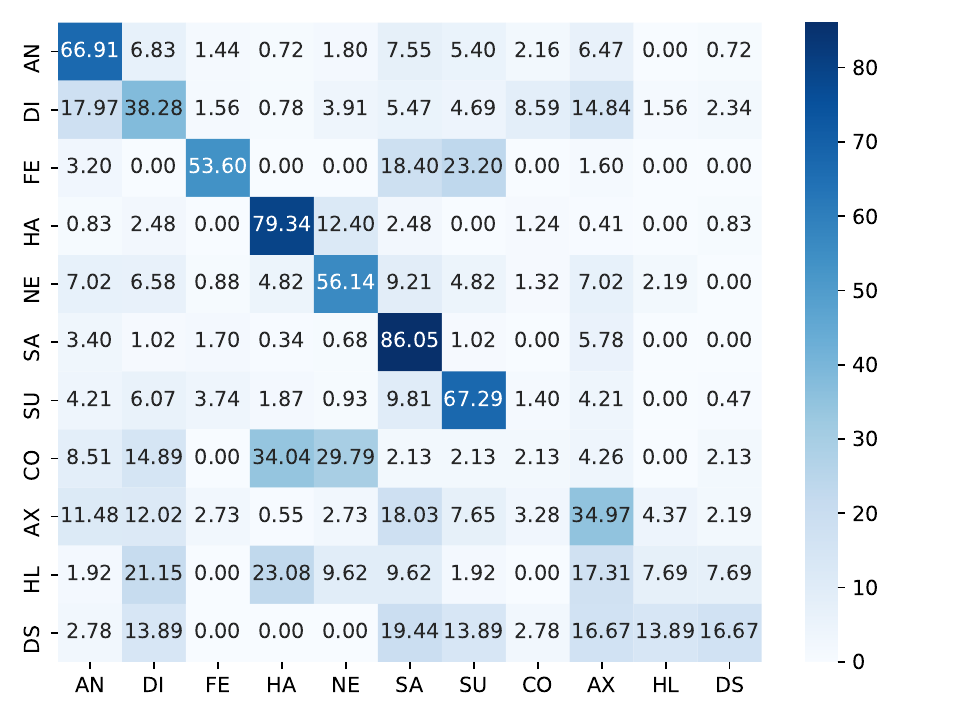}}
	\subfloat[Fold 4\label{fig_conf_mat_mafw_fold_4}]{
		\includegraphics[scale=0.35]{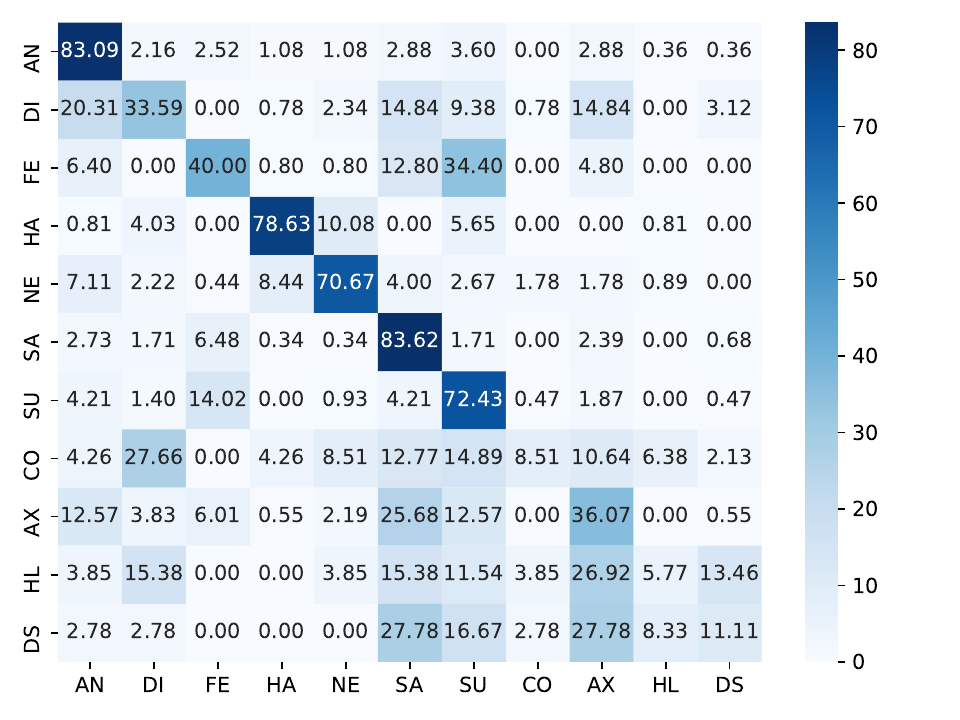}}
	\subfloat[Fold 5\label{fig_conf_mat_mafw_fold_5}]{
		\includegraphics[scale=0.33]{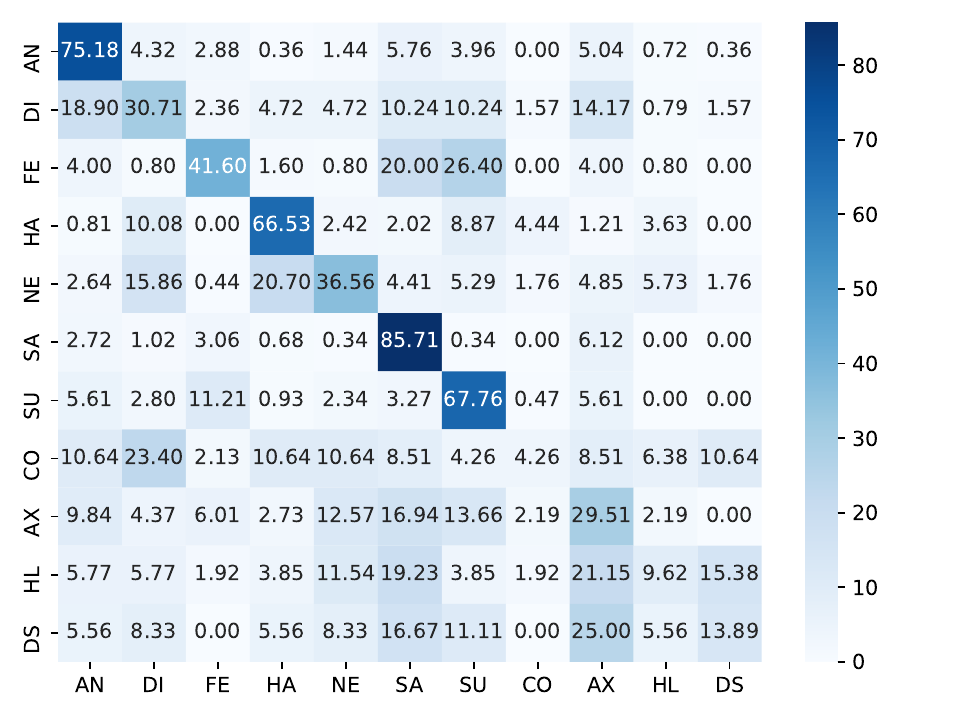}}
	\caption{Confusion matrices on MAFW (11-class). AN: anger. DI: disgust. FE: fear. HA: happiness. NE: neutral. SA: sadness. SU: surprise. CO: contempt. AX: anxiety. HL: helplessness. DS: disappointment.}
	\label{fig_conf_mat} 
\end{figure*}

\subsubsection{Model Scales}
Finally, we show the impact of model scales on downstream fine-tuning. We present the results in Fig. \ref{fig_ablation_model_scales}. Although these numbers are shown in previous tables, we want to offer a more clear and intuitive understanding. As shown in Fig. \ref{fig_ablation_model_scales}, we observe that the larger model typically beats the smaller one in downstream fine-tuning. Moreover, we notice that the performance saturation is less pronounced than that shown in pre-training epochs (especially for MAFW), indicating that further scaling up the model size can yield better results. Thus, we also encourage other researchers to continue this exploration.

\subsection{Visualization Analysis}

\subsubsection{Masked Audio-Visual Reconstruction}
\label{sec_exp_reconstruction_visualization}

We first show the reconstruction ability of HiCMAE under different masking rates. We randomly select video clips of four celebrities from the unseen \textit{test} set of VoxCeleb2 for evaluation. The results of masked audio-visual reconstruction are shown in Fig. \ref{fig_masked_reconstruction}. 
For each video clip, we display the original audio spectrogram and 8 facial frame images in the first row, the masked input in medium level (60\% for audio and 75\% for video) along with the corresponding reconstructed data in the second and third row, and the highly masked input (80\% for audio and 90\% for video) along with the corresponding reconstructed data in the last two rows. From the figure, we observe that, although some fine-grained details are lost, the global structures of the audio spectrogram (e.g., harmonics) and visual frames (e.g., smiles) are well recovered under both medium and high masking rates. 
These satisfactory reconstruction results indicate that HiCMAE is capable of fully exploiting \textit{limited} intra-modal and cross-modal contextual information to infer the missing audio-visual data. We believe this capability learned during self-supervised pre-training has laid the foundation for its superior downstream fine-tuning performance.

\subsubsection{Embedding Space}
In this part, we utilize t-SNE \cite{van2008visualizing} to visualize the learned feature embedding space of HiCMAE on CREMA-D (6-class). To demonstrate the effect of audio-visual fusion, we present both unimodal and multimodal embedding space. The results are shown in Fig. \ref{fig_embedding_space}.
Each row presents unimodal or audio-visual embedding space and each column denotes one of five folds on this dataset. 
From the figure, we find that both audio-only and video-only models have learned good embedding space for distinguishing different kinds of emotions. Moreover, the multimodal embedding space is more discriminative than the unimodal ones, as evidenced by its more compact intra-class and more separated inter-class distributions. Therefore, this comparison result qualitatively verifies the effectiveness of multimodal fusion for audio-visual emotion recognition.

\subsection{Error Analysis}

We present the confusion matrices on MAFW (11-class) in Fig.~\ref{fig_conf_mat}. The aggregated confusion matrix across five folds is shown in Fig.~\ref{fig_conf_mat_mafw} and the confusion matrix of each fold is shown in Fig. \ref{fig_conf_mat_mafw_fold_1}-\ref{fig_conf_mat_mafw_fold_5}. 
As can be seen, although achieving significant improvement over previous methods as stated in Section \ref{sec_exp_mafw_sota}, the performance of HiCMAE on several rare emotions (such as contempt, disappointment, and disgust) is not satisfactory. 
This is mainly attributed to the imbalanced emotion distribution in real-world scenarios. The imbalanced learning strategies (e.g., resampling techniques and cost-sensitive learning) can be utilized to address this problem and we leave it in future work. We also notice that, among all emotions, neutral emotion is the one most easily confused with other emotions.
We believe that the reason is that the boundary between neutral emotion and other emotions is typically less clear than those between other emotion combinations (e.g., happiness and anger). Besides, we observe that fear is often misclassified as surprise. This is consistent with our expectation as their difference in terms of facial expressions is very subtle (both featuring wide-open eyes and raised eyebrows).

\section{Conclusion}
In this paper, we have presented a novel self-supervised framework (HiCMAE), as an early attempt to leverage large-scale self-supervised pre-training to address the dilemma faced by current supervised methods and largely promote the development of AVER. 
HiCMAE is built on top of two primary forms of self-supervision, namely masked data modeling and contrastive learning. Moreover, to facilitate hierarchical audio-visual feature learning, it introduces a three-pronged approach, including hierarchical skip connections between the encoder and decoder, hierarchical cross-modal contrastive learning, and hierarchical feature fusion for downstream fine-tuning. 
Comprehensive experiments across 9 AVER datasets covering both categorical and dimensional tasks demonstrate that HiCMAE outperforms state-of-the-art audio-visual methods by significant margins, indicating that HiCMAE is a powerful audio-visual emotion representation learner. 
Extensive ablation studies and visualization analysis also verify the efficacy of HiCMAE.
We hope this work can provide some insight into the development of AVER and inspire more relevant studies. 


It should be noted that, due to limited computational resources, we cannot afford the pre-training of larger models with more training time and data. Therefore, we plan to tackle these issues in future work and also encourage other researchers to conduct follow-up studies. 
Besides, although HiCMAE has achieved exceptional performance on many datasets, this paper has conducted limited exploration into its internal mechanisms. Therefore, it is worth investigating and enhancing the interpretability of HiCMAE in future work. Possible interpretation tools include attention visualization \cite{dosovitskiy2020image}, relevancy map \cite{chefer2021transformer}, and Grad-CAM \cite{selvaraju2017grad}.
Finally, it is also interesting to apply HiCMAE to other audio-visual tasks (e.g., active speaker detection, deepfake detection, and talking face generation).

\section*{Acknowledgements}
This work is supported by the National Natural Science Foundation of China (NSFC) (No.61831022, No.62276259, No.62201572, No.U21B2010), Beijing Municipal Science \& Technology Commission, Administrative Commission of Zhongguancun Science Park (No.Z211100004821013), Open Research Projects of Zhejiang Lab (No.2021KH0AB06), and CCF-Baidu Open Fund (No.OF2022025).



\bibliographystyle{elsarticle-num} 
\bibliography{main}






\end{document}